\documentclass[runningheads]{llncs}

\usepackage{eccv}

\usepackage{graphicx}
\usepackage{booktabs}

\usepackage[accsupp]{axessibility}  % Improves PDF readability for those with disabilities.

\usepackage{bm}
\usepackage{multirow}
\usepackage{adjustbox}
\usepackage{makecell}

\usepackage[breaklinks,colorlinks]{hyperref}  % CUSTOM: average of eccv22

\usepackage{orcidlink}

\begin{document}

% ---------------------------------------------------------------
\title{
Layer-Wise Relevance Propagation with Conservation Property for ResNet
}

\newcommand*\samethanks[1][\value{footnote}]{\footnotemark[#1]}
\author{
Seitaro Otsuki\inst{1}\orcidlink{0009-0009-8071-6060} \and
Tsumugi Iida\inst{1} \and
F\'elix Doublet\inst{1} \and
Tsubasa Hirakawa\inst{2}\orcidlink{0000-0003-3851-5221} \and
Takayoshi Yamashita\inst{2}\orcidlink{0000-0003-2631-9856} \and
Hironobu Fujiyoshi\inst{2}\orcidlink{0000-0001-7391-4725} \and
Komei Sugiura\inst{1}\orcidlink{0000-0002-0261-0510}
}

\authorrunning{S.~Otsuki et al.}

\institute{
Keio University, Japan
\\\email{\{otsu8sei14, tiida, felixdoublet, komei.sugiura\}@keio.jp}
\and
Chubu University, Japan
\\\email{hirakawa@mprg.cs.chubu.ac.jp, \{takayoshi, fujiyoshi\}@isc.chubu.ac.jp}
}

\maketitle

\begin{abstract}
  % #line of eccv22 papers: 19, 21, 16, 20, 19 ==> ave.: 19
  % following has 17 lines (ave. - 2)
  The transparent formulation of explanation methods is essential for elucidating the predictions of neural networks, which are typically black-box models. Layer-wise Relevance Propagation (LRP) is a well-established method that transparently traces the flow of a model's prediction backward through its architecture by backpropagating relevance scores.
  However, the conventional LRP does not fully consider the existence of skip connections, and thus its application to the widely used ResNet architecture has not been thoroughly explored.
  In this study, we extend LRP to ResNet models by introducing Relevance Splitting at points where the output from a skip connection converges with that from a residual block. Our formulation guarantees the conservation property throughout the process, thereby preserving the integrity of the generated explanations.
  To evaluate the effectiveness of our approach, we conduct experiments on ImageNet and the Caltech-UCSD Birds-200-2011 dataset.
  Our method achieves superior performance to that of baseline methods on standard evaluation metrics such as the Insertion-Deletion score while maintaining its conservation property.
  We will release our code for further research at \url{https://5ei74r0.github.io/lrp-for-resnet.page/}
\end{abstract}

\section{Introduction}
\label{sec:intro}
The widespread adoption of neural networks underscores the critical importance of explainability of these models~\cite{shrikumar17deeplift,ribeiro2016lime}.
Indeed, the European Parliament's AI Act, promulgated in December 2023, proclaims that AI systems must be safe and transparent~\cite{aiact}.
This strengthens the need to develop the methods to generate appropriate and meaningful explanations of neural network models.
Current methods often lack transparency, making the interpretation of the results a non-trivial task~\cite{molnar2020interpretable}.
Additionally, the black-box nature of neural network models sometimes masks the underlying logic of their inference processes. This opacity presents significant challenges in verifying the validity of the models' predictions.
To address this issue, Explainable AI has been proposed as a means of making the models more transparent and thus promoting the application of AI in critical domains~\cite{SAEED2023110273}.

The generation of visual explanations within neural networks poses a significant challenge, necessitating the precise extraction of critical areas.
For example, the current Layer-wise Relevance Propagation (LRP)~\cite{bach2015layerwiserelevancepropagationlrp} implementation encounters notable challenges when applied to ResNet~\cite{he2016deep} architectures. 
ResNet's residual connections create multiple non linear relevance pathways, which cannot be handled by the typical relevance attribution process of LRP.
Consequently, the explanations generated by LRP for ResNet models often give unreliable results as shown in \cref{sec:qualitative}.
Therefore, while LRP offers a valuable framework for relevance attribution in neural networks, its limitations, particularly in handling architectures like ResNet, pose a challenge. This underscores the pressing need for advanced methodologies in this domain.

\begin{figure}[!t]
    \centering
    \includegraphics[width=\linewidth]{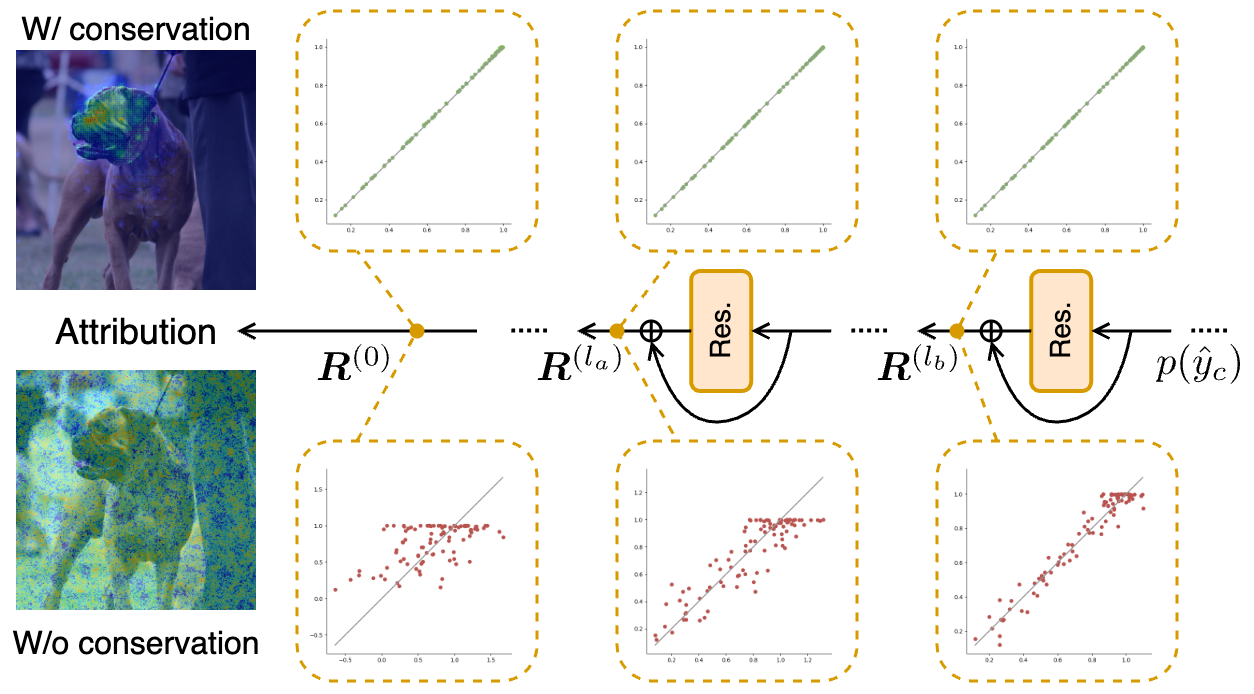}
    \caption{
    We propose LRP for ResNet. 
    By formulating Relevance Splitting at a point where the output from a skip connection converges with that from a residual block, we extend LRP---originally designed for propagating relevance between two consecutive layers---to the ResNet architecture while guaranteeing its conservation property, thereby preserving the integrity of the explanation process.
    }
    \label{fig:eye-catch}
\end{figure}

In explaining inference processes of black-box models, it is crucial that the explanation method is transparently formulated. 
Although several explanation methods applicable to ResNet, such as Grad-CAM~\cite{Selvaraju2017gradcam}, RISE~\cite{Petsiuk2018rise}, and LIME~\cite{ribeiro2016lime}, offer valuable insights, they often lack transparency. For instance, Grad-CAM relies on the output of the last convolutional layer for its explanation, which may obscure the contributions of earlier model structures. Although Grad-CAM effectively highlights relevant areas in the input image for the model's inference, it falls short in elucidating how the earlier layers and their interconnections influence these findings. This limitation prevents a comprehensive understanding of the model’s internal inference mechanisms.

By contrast, LRP~\cite{bach2015layerwiserelevancepropagationlrp} is a transparent explanation method that traces the flow of a model's prediction back through its architecture by backpropagating relevance scores.
A key aspect of LRP is its conservation property, which ensures that the total relevance score received by an unit is equal to the total it redistributes.
However, the application of LRP to CNN models has been limited. Although it has been applied to traditional models such as VGG~\cite{simonyan2015vgg}, LRP's application to the widely used~\cite{ren2017faster,DiffusionDet,detic,kamath2021mdetr,radford2021learning,NEURIPS2023_c1f7b1ed,A-Generalist-Agent}
ResNet architecture has not been sufficiently discussed.

In this study, we extend LRP to models with residual connections, such as ResNet, by formulating a novel relevance propagation rule that maintains the conservation property. This adaptation allows high-quality, transparent explanations to be generated for ResNet models, thereby contributing to a deeper understanding and improved interpretability of complex neural networks.
A comprehensive summary of our method is shown in \cref{fig:eye-catch}.

Our method is potentially applicable to other models featuring residual connections.
Previous studies~\cite{arras2017explaininglrplstm,ali2022lrptransformer} have applied LRP to models such as LSTM~\cite{Hochreiter1997LSTM} and transformers~\cite{vaswani2017attention}. However, these adaptations have not addressed relevance propagation in the presence of residual connections. Specifically, the application of LRP to transformers has ignored the existence of residual connections in its relevance propagation~\cite{ali2022lrptransformer}.
Our key contributions are as follows:
\begin{itemize}
    \item [$\bullet$] We extend LRP to models with residual connections by introducing Relevance Splitting at points where the output from a skip connection converges with that from a residual block.
    \item [$\bullet$] The conservation property is guaranteed throughout the proposed process, thereby preserving the integrity of the relevance propagation mechanism.
    \item [$\bullet$] To mitigate the issue of overconcentration of generated attributions within irrelevant regions, we introduce Heat Quantization.
    \item [$\bullet$] Our method demonstrate superior performance to that of baseline methods on standard evaluation metrics, including the Insertion-Deletion score.
    \item [$\bullet$] We investigate improved designs for relevance propagation within residual blocks and skip connections through an ablation study
    % on relevance propagation within the Bottleneck module 
    and demonstrate the significance of propagating relevance through skip connections and employing Ratio-Based Relevance Splitting.
\end{itemize}

\section{Related Work}
\label{sec:related}
There has been widespread research on generating visual explanations for neural network models~\cite{pan2021iared,bach2015layerwiserelevancepropagationlrp,binderLayerwiseRelevancePropagation2016,cheferTransformerInterpretabilityAttention2021,Selvaraju2017gradcam,wangScoreCAMScoreWeightedVisual2020,fukui2019attention,jacovi-etal-2023-neighboring}.
General post-hoc explanation methods are primarily categorized into three types based on the method whereby they generate explanations: perturbation, backpropagation, and approximation.

\subsubsection{Perturbation methods.}
Perturbation methods such as RISE~\cite{Petsiuk2018rise}, extremal perturbations~\cite{fong2017interpretable}, and SHAP~\cite{lundberg2017shap} generate explanations by deliberately modifying the input images. These methods are effective, but often entail multiple interactions with black-box models, resulting in time-consuming processes that pose a significant drawback for practical applications.

\subsubsection{Backpropagation methods.}
Backpropagation methods~\cite{simonyan2014deep,shrikumar2016gradientxinput,sundararajan2017integratedgrad,shrikumar17deeplift,bach2015layerwiserelevancepropagationlrp,Selvaraju2017gradcam,srinivas2019fullgrad} leverage the backpropagation algorithm to produce gradient or gradient-related explanations.
Integrated Gradient~\cite{sundararajan2017integratedgrad} utilizes the gradient theorem to allocate attributions to input features.
Grad-CAM~\cite{Selvaraju2017gradcam} calculates explanations by summing CNN activations across channels, with weights derived from the average of the corresponding gradients. Rule-based backpropagation methods such as DeepLift~\cite{shrikumar17deeplift} and Layer-wise Relevance Propagation (LRP)~\cite{bach2015layerwiserelevancepropagationlrp}, which propagate scores using distinct methodologies. LRP is a well-established method that guarantees the conservation property; however, its proper application for ResNet is currently limited.
% [added after the rebuttal]
Although the original LRP is applicable to simple skip connections with identity weights, we cannot directly apply the original LRP to skip connections with upscaling or downscaling mappings.

In this study, we integrates LRP
into the ResNet architecture. This integration marks a significant advance over the traditional application of LRP, which lacks support for models with residual connections.

\subsubsection{Approximation methods.}
Approximation methods~\cite{ribeiro2016lime,parekh2021framework} employ an external entity to elucidate inference processes of black-box models.
These methods create understandable approximations or explanations of how the black-box model works, usually focusing on specific predictions or decisions.
However, their separation from the black-box models can pose challenges in accurately capturing the models' intricate working.

\subsubsection{Incorporating modules for explanation.}
Another body of research~\cite{fukui2019attention,OguraMSHYFK20aben,itaya2021maska3cexplanationattentionbranch,Iida2022lambdaattetnionbranchnetworkslabn} has investigated the direct incorporation of modules designed for generating explanations into the model architecture.
For example, the Attention Branch Network~\cite{fukui2019attention} enhances image recognition performance by simultaneously generating explanations. However, the integration of such explanation-centric modules adds a layer of complexity, potentially reducing the model's overall transparency.

\subsubsection{Datasets.}
In the pursuit of generating visual explanations for image classification tasks, standard datasets such as ImageNet~\cite{deng2009Imagenet}, CIFAR-10 and CIFAR-100~\cite{krizhevsky2009learning} are commonly employed. In addition to these general datasets, domain-specific datasets such as Caltech-UCSD Birds-200-2011~\cite{WahCUB_200_2011} and the Indian Diabetic Retinopathy Image Dataset (IDRiD)~\cite{PORWALidrid} are often used. CUB features annotated bird images from 200 species, along with 15 part attributes per species. IDRiD contains fundus images with annotations of varying severity levels of fundus hemorrhage as evaluated by medical professionals.

\begin{figure}[!t]
    \centering
    \includegraphics[width=\linewidth]{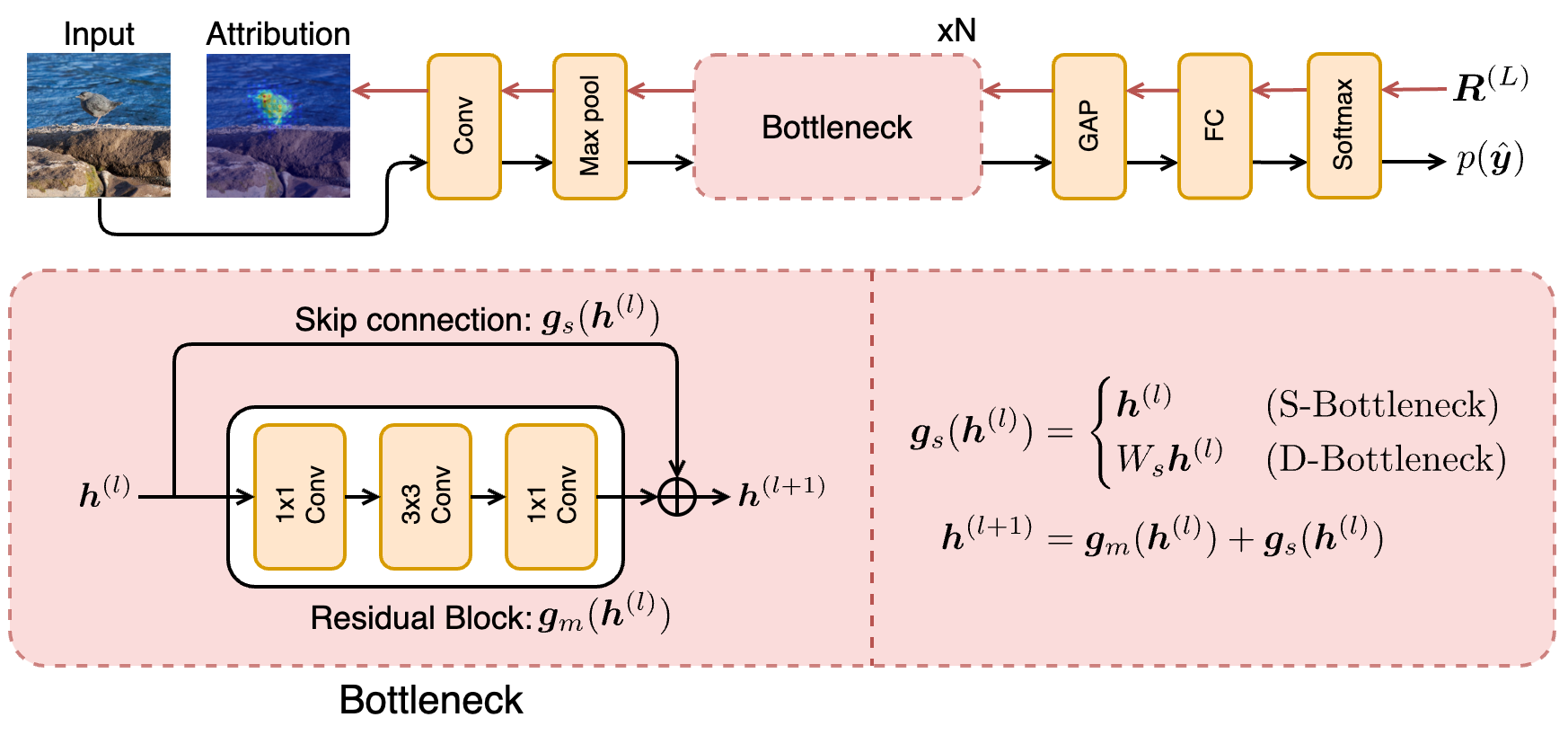}
    \caption{LRP for ResNet. \textbf{Top:} LRP propagates the relevance score backward to generate an attribution map corresponding to the input image. We focus on the relevance score propagation through the Bottleneck module, which incorporates a residual connection. \textbf{Bottom:} Architecture of the Bottleneck module. The D-Bottleneck employs a linear projection,
    in its skip connection for dimension matching. ReLU activation functions and batch normalization layers are omitted for simplicity.}
    \label{fig:framework}
\end{figure}

\section{Background: Layer-wise Relevance Propagation}
In the following sections, we first briefly review Layer-wise Relevance Propagation (LRP)~\cite{bach2015layerwiserelevancepropagationlrp} before discussing how it should be extended to ResNet and introducing several relevance propagation rules.

LRP is an explanation method that propagates a model's prediction back through the network via relevance scores. Its key feature is the propagation property, ensuring that the sum of relevance scores received by an unit equals the sum redistributed by the same unit.

\subsection{LRP Between Two Consecutive Layers}
Consider $\bm h^{(l)} \in \mathbb{R}^D$ and $\bm h^{(l+1)} \in \mathbb{R}^E$ as the intermediate features in the $l$-th and $(l+1)$-th consecutive layers, respectively. Let $\bm f_{\bm \theta}$ be the function parametrized by $\bm \theta$ which maps $\bm h^{(l)}$ to $\bm h^{(l+1)}$, and define $\bm R^{(l)}$ and $z_{i j}$ as the relevance scores of $\bm h^{(l)}$ and the contribution from $h^{(l)}_i$ to $h^{(l+1)}_j$, respectively.

The relevance scores propagating through $\bm f_{\bm \theta}$ from all $h^{(l+1)}_j$ to $h^{(l)}_i$ can then be expressed as:
\begin{align}
R^{(l)}_i = \sum^{E}_{j=1} \frac{z_{i j}}{\sum^D_{k=1} z_{k j}} R^{(l+1)}_j, 
\label{equ:lrp}
\end{align}
where $R^{(l+1)}_j$ and $R^{(l)}_i$ denote the relevance scores of the $j$-th element of $\bm h^{(l+1)}$ and the $i$-th element of $\bm h^{(l)}$, respectively. Here, the quantity $z_{i j}$ can be formulated in various ways.
The denominator $\sum^D_{k=1} z_{k j}$ ensures the conservation of the total relevance score between two consecutive layers.

In this study, we employ the $z^+$-Rule~\cite{montavon2017dtd} for the relevance propagation where we can view $\bm f_{\bm \theta}$ as a linear projection.
The relevance propagation through linear projection $\bm{f}$ in the $z^+$-Rule can be written as:
\begin{align}
R^{(l)}_i &= \sum^{E}_{j=1} \frac{w^+_{j i} h^{(l)}_i}{\sum^D_{k=1} w^+_{j k} h^{(l)}_k} R^{(l+1)}_j,
\end{align}
where $\bm{f} (\bm h^{(l)}) = W \bm h^{(l)}$, $w^+_{j i} = \max(0, w_{j i})$, and $w_{j i}$ denotes the $(j, i)$-th element of $W \in \mathbb{R}^{E \times D}$. In this context, $W^+$ is defined as the matrix consisting of the non-negative elements of $W$, that is, $W^+_{i j} = \max(0, W_{i j})$ for all $i$ and $j$. This allows us to reformulate the equation as:
\begin{align}
R^{(l)}_i = \sum^{E}_{j=1} \frac{\frac{\partial f^+_j}{\partial h^{(l)}_i}(\bm h^{(l)}) h^{(l)}_i}{f^+_j (\bm h^{(l)})} R^{(l+1)}_j,
\end{align}
where $\bm{f}^+ (\bm h^{(l)}) = W^+ \bm h^{(l)}$.

We apply this rule to convolution, global average pooling (GAP), max pooling, and fully-connected layers, as these operations can be formulated as linear projections with specific weight matrices. For ReLU activation functions and batch normalization (BN) layers, the relevance scores are passed through without any modifications.

\section{Method}
\label{sec:method}
By introducing a novel relevance propagation rule for models with residual connections, we extend Layer-wise Relevance Propagation (LRP)~\cite{bach2015layerwiserelevancepropagationlrp} to be applicable in such models, specifically developing an LRP for ResNet~\cite{he2016deep} models.
This extension
defines the calculation method of LRP for models with residual connections. Consequently, our approach is widely applicable to models possessing residual blocks.
\cref{fig:framework} provides an overview of the ResNet architecture and the associated application of LRP.

\subsection{General Formulation}
\begin{figure}[!t]
    \centering

    % ENTIRE BOX
    \begin{minipage}{0.60\linewidth}
        \hfill
        \begin{minipage}{0.42\linewidth}
            \centering
            \raisebox{5pt}{\includegraphics[width=\linewidth]{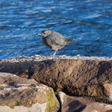}}
        \end{minipage}
        \hfill
        \begin{minipage}{0.1\linewidth}
        \end{minipage}
        \hfill
        \begin{minipage}{0.42\linewidth}
            \centering
            \raisebox{5pt}{\includegraphics[width=\linewidth]{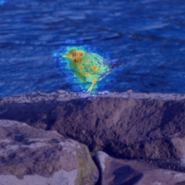}}
        \end{minipage}
        \hfill

        % Row 2
        \hfill
        \begin{minipage}[t]{0.42\linewidth}
            \centering ``Water ouzel''
        \end{minipage}
        \hfill
        \begin{minipage}{0.1\linewidth}
        \end{minipage}
        \hfill
        \begin{minipage}[t]{0.42\linewidth}
            \centering Relevant attribution
        \end{minipage}
        \hfill
    \end{minipage}
    \caption{\textbf{Left:} Typical sample of an input image from ImageNet~\cite{deng2009Imagenet}. \textbf{Right:} the corresponding attribution as a visual explanation.}
    \label{fig:example}
\end{figure}

In this study, we focus on the task of visualizing important regions in an image as a visual explanation of the model's prediction. In this task, the visual explanation should focus on the pixels that contributed to the model's decision.
\cref{fig:example} shows a typical sample of the task. The left and right panels show the input image and the visual explanation, respectively.
The input is an image $\bm{x} \in \mathbb{R}^{c^{(0)} \times h^{(0)} \times w^{(0)}}$, where $c^{(0)}, h^{(0)}$, and $w^{(0)}$ denote the number of channels, height and width of the input image, respectively.
The output $p(\hat{\bm y}) \in [0, 1]^C$ denotes the predicted probability for each class, where $C$ is the number of classes.
Additionally, the importance of each pixel is obtained as an attribution $\bm{\alpha} \in  \mathbb{R}^{h^{(0)} \times w^{(0)}}$ which is used as a visual explanation.
To quantitatively evaluate our method, we use the Insertion, Deletion, and Insertion-Deletion scores~\cite{Petsiuk2018rise}. In this study, we assume that the model is based on a ResNet architecture.

\subsection{Architecture of ResNet50}
\label{sec:Architecture-of-ResNet50}
Before discussing the extension of LRP to ResNet models, we briefly review the architecture of ResNet, with a focus on ResNet50. % added in nagashi-komi
ResNet50 consists of a convolution layer, a BN layer, a max pooling layer, 16 Bottleneck modules, a GAP layer, and a fully-connected layer followed by a softmax function. Each Bottleneck module consists of a skip connection and a residual block. The residual block is structured with three convolution layers: a 1 $\times$ 1 convolution for dimension reduction, a 3 $\times$ 3 convolution for processing, and another 1 $\times$ 1 convolution to restore or increase the dimension. These layers are sequentially followed by BN and ReLU activation functions. 

We treat two types of Bottleneck modules: the Simple Bottleneck (S-Bottlene\-ck) module and the Downsampling Bottleneck (D-Bottleneck) module. The S-Bottleneck module uses an identity mapping in its skip connection. In contrast, the D-Bottleneck module employs a linear projection in its skip connection for dimension matching.
In this study, we mainly focus on how relevance should be propagated through these Bottleneck modules.
% WARN: above sentence uses `\-`.

\subsection{LRP for Bottleneck Modules}
\label{sec:LRP-for-the-Bottleneck-modules}
In this section, we discuss and formulate several potential relevance propagation rules for Bottleneck modules. As seen in \cref{equ:lrp}, LRP originally defines a propagation rule between two consecutive layers. However, this rule does not account for the existence of a skip connection that bridges nonconsecutive layers.
This raises a fundamental question: how should we propagate the relevance scores at the point where the output of the skip connection converges with that of the residual block?

As a preliminary step, we first divide the relevance score $ \bm R^{(l+1)} $ into two parts: $ \bm R_s $ for propagation through the skip connection and $ \bm R_m $ for the mainstream of the residual block. In this context, $ s $ in $ \bm R_s $ signifies the skip connection, and $ m $ in $ \bm R_m $ denotes the mainstream of the residual block. To adhere to the conservation property,
we impose the constraint: \(\bm R^{(l)} = \bm R^{(l+1)} = \bm R_s + \bm R_m.\)

\begin{figure}[!t]
    \centering
    \includegraphics[width=\linewidth]{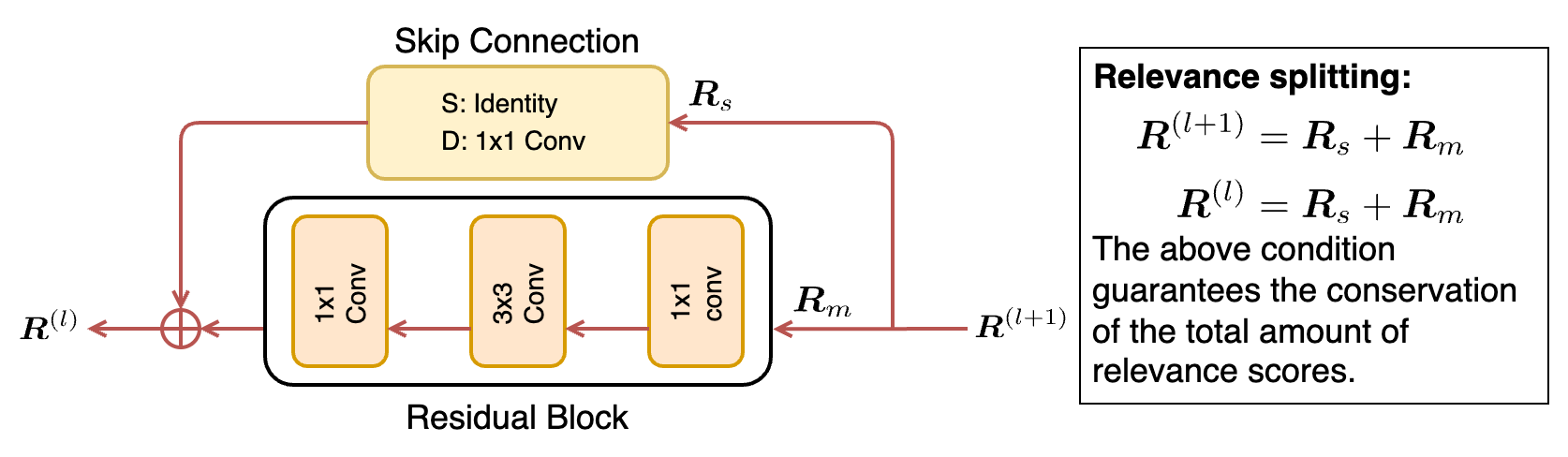}
    \caption{Architecture of the Bottleneck block in ResNet and our Relevance Splitting approach. We introduce Relevance Splitting to consider the existence of skip connections in the relevance propagation of LRP.
    }
    \label{fig:relevance}
\end{figure}

We formulate two approaches for dividing $ \bm R_s $ and $ \bm R_m $: Symmetric Splitting and Ratio-Based Splitting, as illustrated in \cref{fig:relevance}.
Symmetric Splitting divides $ \bm R^{(l+1)} $ equally as follows:
\begin{align}
(R_s)_i = (R_m)_i = \frac{R^{(l+1)}_i}{2}.
\end{align}
This method is straightforward, but does not account for the varying contributions of individual elements in the intermediate feature $ \bm h^{(l+1)} $, which could lead to a less-nuanced propagation of relevance.

Ratio-Based Splitting is a more thorough approach. Let $ \bm h_s $ and $ \bm h_m $ represent the outputs of the skip connection and the residual block, respectively. We divide $ \bm R^{(l+1)} $ to satisfy the following conditions:
\begin{align}
(R_s)_i = \frac{R^{(l+1)}_i \cdot |(h_s)_i|}{|(h_m)_i| + |(h_s)_i|},\quad (R_m)_i = \frac{R^{(l+1)}_i \cdot |(h_m)_i|}{|(h_m)_i| + |(h_s)_i|}.
\end{align}
This approach accounts for the ratio of the absolute values of $ \bm h_s $ and $ \bm h_m $. When the skip connection is an identity mapping, $ \bm h_s = \bm h^{(l)} $, and $ \bm h_m $ is equivalent to $ \bm h^{(l+1)} - \bm h^{(l)} $, representing the change in features attributable to the model's parameters. In such cases, the greater the absolute value of an element in $ \bm h_m $, the more significant the model's contribution to that element.
This approach naturally satisfies the conservation property:
\begin{align}
   (R_s)_i + (R_m)_i &= \frac{R^{(l+1)}_i \cdot |(h_s)_i|}{|(h_m)_i| + |(h_s)_i|} + \frac{R^{(l+1)}_i \cdot |(h_m)_i|}{|(h_m)_i| + |(h_s)_i|} \\&= \frac{R^{(l+1)}_i \cdot (|(h_m)_i| + |(h_s)_i|)}{|(h_m)_i| + |(h_s)_i|} = R^{(l+1)}_i. 
\end{align}

\subsection{LRP for Two Types of Skip Connections}
\label{sec:LRP-for-two-types-of-skip-connections}
As mentioned in \cref{sec:Architecture-of-ResNet50}, there are two types of Bottleneck modules, each with a distinct type of skip connection. Specifically, the S-Bottleneck module features an identity mapping in its skip connection, whereas the D-Bottleneck module uses a linear projection.

This distinction leads to another question: should the two types of skip connections, i.e., identity mapping and linear projection, be treated equally?
It appears logical to propagate relevance scores through linear projections in skip connections, as they transform the input through multiplication with their parameter matrices.
These operations in the D-Bottleneck modules, although primarily intended to downsample the input to match the output dimension, can selectively emphasize important input elements through a linear projection.
In contrast, a skip connection with an identity mapping does not perform any transformation.
This suggests that there might be room to apply different relevance propagation approaches for the two types of skip connections.
One approach entails setting $ \bm R_s = 0 $ for skip connections with identity mappings, thereby propagating all relevance through the residual block. The other approach involves applying the Relevance Splitting method discussed in \cref{sec:LRP-for-the-Bottleneck-modules}, even to skip connections with identity mappings.

Based on the results of preliminary experiments, our proposed explanation method employs Ratio-Based Splitting and applies it to all skip connections, including those with identity mappings. To evaluate the effectiveness of this strategy, we conduct ablation studies comparing these conditions (see \cref{sec:ablation}).

By sequentially applying the described propagation rules and backpropagating the relevance scores, we can compute the relevance score $\bm R^{(0)}$ for the input $\bm x$. Similar to existing LRP methods, $\bm R^{(0)}$ has the same shape as $\bm x$, and its channel-wise sum, denoted as $\bm \alpha_R$, can be directly used as an attribution map. However, $\bm \alpha_R$ tends to excessively concentrate on 
irrelevant regions.
To mitigate this tendency, we quantize the values in $\bm \alpha_R$ to obtain a more even distribution of attribution, resulting in the final attribution map $\bm \alpha$. We refer to this operation as Heat Quantization,
formulated as follows:
\begin{align}
    \alpha_{i,j} &= (\alpha_R)_\mathrm{min} + \left\lfloor \frac{(\alpha_R)_{i,j} - (\alpha_R)_\mathrm{min}}{\left((\alpha_R)_\mathrm{max} - (\alpha_R)_\mathrm{min}\right) \slash\: Q} \right\rfloor Q,
\end{align}
where $Q$ denotes the number of quantization bins. We set $Q=8$.

\begin{figure*}[!t]
    \centering
    % Row 1
    \begin{minipage}{0.13\linewidth}
        \raisebox{2pt}{\includegraphics[width=\linewidth]{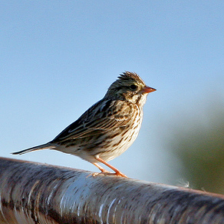}}
    \end{minipage}
    \hfill
    \begin{minipage}{0.13\linewidth}
        \raisebox{2pt}{\includegraphics[width=\linewidth]{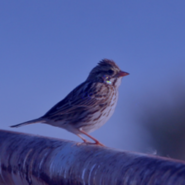}}
    \end{minipage}
    \hfill
    \begin{minipage}{0.13\linewidth}
        \raisebox{2pt}{\includegraphics[width=\linewidth]{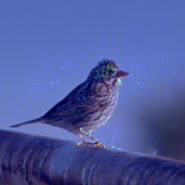}}
    \end{minipage}
    \hfill
    \begin{minipage}{0.13\linewidth}
        \raisebox{2pt}{\includegraphics[width=\linewidth]{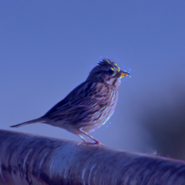}}
    \end{minipage}
    \hfill
    \begin{minipage}{0.13\linewidth}
        \raisebox{2pt}{\includegraphics[width=\linewidth]{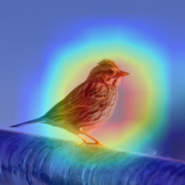}}
    \end{minipage}
    \hfill
    \begin{minipage}{0.13\linewidth}
        \raisebox{2pt}{\includegraphics[width=\linewidth]{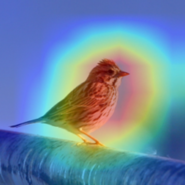}}
    \end{minipage}
    \hfill
    \begin{minipage}{0.13\linewidth}
        \raisebox{2pt}{\includegraphics[width=\linewidth]{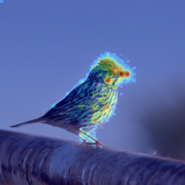}}
    \end{minipage}

    % Row 2
    \begin{minipage}{0.13\linewidth}
        \raisebox{2pt}{\includegraphics[width=\linewidth]{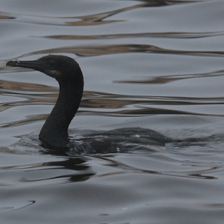}}
    \end{minipage}
    \hfill
    \begin{minipage}{0.13\linewidth}
        \raisebox{2pt}{\includegraphics[width=\linewidth]{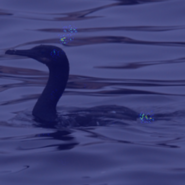}}
    \end{minipage}
    \hfill
    \begin{minipage}{0.13\linewidth}
        \raisebox{2pt}{\includegraphics[width=\linewidth]{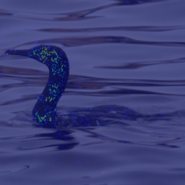}}
    \end{minipage}
    \hfill
    \begin{minipage}{0.13\linewidth}
        \raisebox{2pt}{\includegraphics[width=\linewidth]{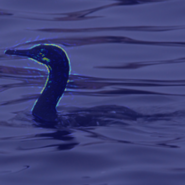}}
    \end{minipage}
    \hfill
    \begin{minipage}{0.13\linewidth}
        \raisebox{2pt}{\includegraphics[width=\linewidth]{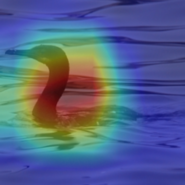}}
    \end{minipage}
    \hfill
    \begin{minipage}{0.13\linewidth}
        \raisebox{2pt}{\includegraphics[width=\linewidth]{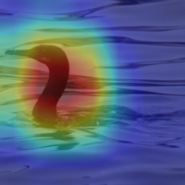}}
    \end{minipage}
    \hfill
    \begin{minipage}{0.13\linewidth}
        \raisebox{2pt}{\includegraphics[width=\linewidth]{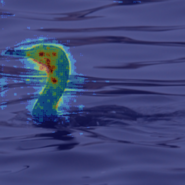}}
    \end{minipage}

    % Row 3
    \begin{minipage}{0.13\linewidth}
        \raisebox{2pt}{\includegraphics[width=\linewidth]{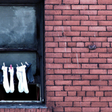}}
    \end{minipage}
    \hfill
    \begin{minipage}{0.13\linewidth}
        \raisebox{2pt}{\includegraphics[width=\linewidth]{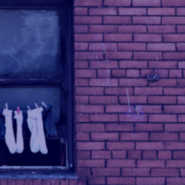}}
    \end{minipage}
    \hfill
    \begin{minipage}{0.13\linewidth}
        \raisebox{2pt}{\includegraphics[width=\linewidth]{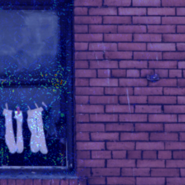}}
    \end{minipage}
    \hfill
    \begin{minipage}{0.13\linewidth}
        \raisebox{2pt}{\includegraphics[width=\linewidth]{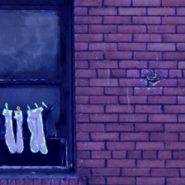}}
    \end{minipage}
    \hfill
    \begin{minipage}{0.13\linewidth}
        \raisebox{2pt}{\includegraphics[width=\linewidth]{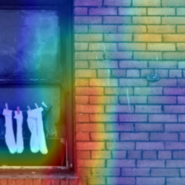}}
    \end{minipage}
    \hfill
    \begin{minipage}{0.13\linewidth}
        \raisebox{2pt}{\includegraphics[width=\linewidth]{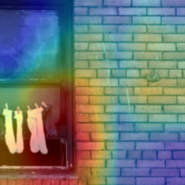}}
    \end{minipage}
    \hfill
    \begin{minipage}{0.13\linewidth}
        \raisebox{2pt}{\includegraphics[width=\linewidth]{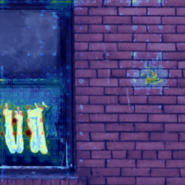}}
    \end{minipage}

    % Row 4
    \begin{minipage}{0.13\linewidth}
        \raisebox{2pt}{\includegraphics[width=\linewidth]{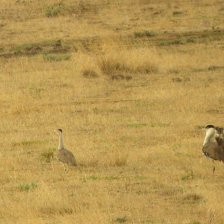}}
    \end{minipage}
    \hfill
    \begin{minipage}{0.13\linewidth}
        \raisebox{2pt}{\includegraphics[width=\linewidth]{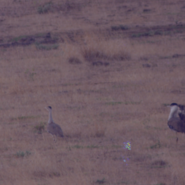}}
    \end{minipage}
    \hfill
    \begin{minipage}{0.13\linewidth}
        \raisebox{2pt}{\includegraphics[width=\linewidth]{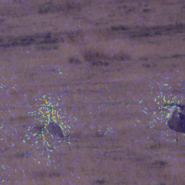}}
    \end{minipage}
    \hfill
    \begin{minipage}{0.13\linewidth}
        \raisebox{2pt}{\includegraphics[width=\linewidth]{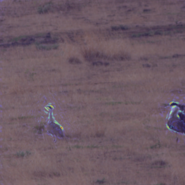}}
    \end{minipage}
    \hfill
    \begin{minipage}{0.13\linewidth}
        \raisebox{2pt}{\includegraphics[width=\linewidth]{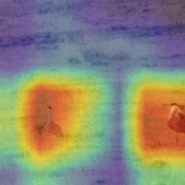}}
    \end{minipage}
    \hfill
    \begin{minipage}{0.13\linewidth}
        \raisebox{2pt}{\includegraphics[width=\linewidth]{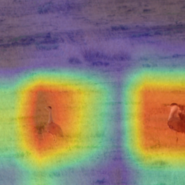}}
    \end{minipage}
    \hfill
    \begin{minipage}{0.13\linewidth}
        \raisebox{2pt}{\includegraphics[width=\linewidth]{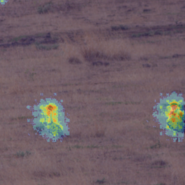}}
    \end{minipage}

    % Row 5
    \begin{minipage}{0.13\linewidth}
        \raisebox{2pt}{\includegraphics[width=\linewidth]{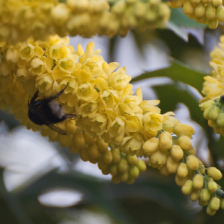}}
    \end{minipage}
    \hfill
    \begin{minipage}{0.13\linewidth}
        \raisebox{2pt}{\includegraphics[width=\linewidth]{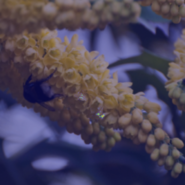}}
    \end{minipage}
    \hfill
    \begin{minipage}{0.13\linewidth}
        \raisebox{2pt}{\includegraphics[width=\linewidth]{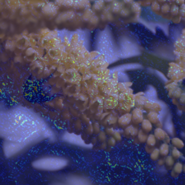}}
    \end{minipage}
    \hfill
    \begin{minipage}{0.13\linewidth}
        \raisebox{2pt}{\includegraphics[width=\linewidth]{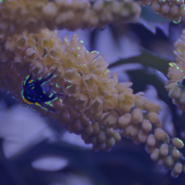}}
    \end{minipage}
    \hfill
    \begin{minipage}{0.13\linewidth}
        \raisebox{2pt}{\includegraphics[width=\linewidth]{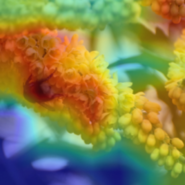}}
    \end{minipage}
    \hfill
    \begin{minipage}{0.13\linewidth}
        \raisebox{2pt}{\includegraphics[width=\linewidth]{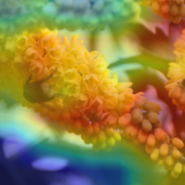}}
    \end{minipage}
    \hfill
    \begin{minipage}{0.13\linewidth}
        \raisebox{2pt}{\includegraphics[width=\linewidth]{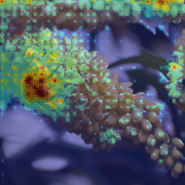}}
    \end{minipage}

    % Row 6
    \begin{minipage}{0.13\linewidth}
        \centering (a) 
    \end{minipage}
    \hfill
    \begin{minipage}{0.13\linewidth}
        \centering (b) 
    \end{minipage}
    \hfill
    \begin{minipage}{0.13\linewidth}
        \centering (c) 
    \end{minipage}
    \hfill
    \begin{minipage}{0.13\linewidth}
        \centering (d)
    \end{minipage}
    \hfill
    \begin{minipage}{0.13\linewidth}
        \centering (e)
    \end{minipage}
    \hfill
    \begin{minipage}{0.13\linewidth}
        \centering (f) 
    \end{minipage}
    \hfill
    \begin{minipage}{0.13\linewidth}
        \centering (g)
    \end{minipage}

    % Row 7
    \begin{minipage}[t]{0.13\linewidth}
        \centering Original
    \end{minipage}
    \hfill
    \begin{minipage}[t]{0.13\linewidth}
        \centering LRP~\cite{bach2015layerwiserelevancepropagationlrp}
    \end{minipage}
    \hfill
    \begin{minipage}[t]{0.13\linewidth}
        \centering IG~\cite{sundararajan2017integratedgrad}%\scalebox{0.95}[1]{\cite{sundararajan2017integratedgrad}}%
    \end{minipage}
    \hfill
    \begin{minipage}[t]{0.13\linewidth}
        \centering \scalebox{0.97}[1]{Guided BP} \cite{springenberg2015striving}
    \end{minipage}
    \hfill
    \begin{minipage}[t]{0.13\linewidth}
        \centering \scalebox{0.97}[1]{Grad-CAM} \cite{Selvaraju2017gradcam}
    \end{minipage}
    \hfill
    \begin{minipage}[t]{0.13\linewidth}
        \centering \scalebox{0.97}[1]{Score-CAM} \cite{wangScoreCAMScoreWeightedVisual2020}
    \end{minipage}
    \hfill
    \begin{minipage}[t]{0.13\linewidth}
        \centering Ours
    \end{minipage}

    \caption{Qualitative Results: Attribution produced by each explanation method for the prediction of ResNet50 with respect to the ground-truth classes (top to bottom):
    ``Brandt Cormorant,''
    ``Savannah Sparrow,''
    ``Sock,''
    ``Bustard,''
    and ``Bee.''
    IG and Guided BP denote Integrated Gradients and Guided BackPropagation, respectively.
    }
    \label{fig:qual}
\end{figure*}

\section{Experiments and Results}
\label{sec:exp}
\subsection{Experimental Setup}
We used the Caltech-UCSD Birds-200-2011 (CUB) dataset~\cite{WahCUB_200_2011} and the validation set of ImageNet~\cite{deng2009Imagenet} (ILSVRC) 2012 to evaluate our method.
These datasets were chosen because they are standard datasets for visual explanation generation tasks.
The CUB dataset contains 11,788 images from 200 classes of bird species.
The validation set of ImageNet consists of 50,000 images from 1,000 classes.

For the experiments on the CUB dataset and ImageNet, we employed ResNet\-50~\cite{he2016deep} that was trained on the CUB dataset and pretrained on ImageNet.
Detailed information about the training setup, data preprocessing, and hardware specifications can be found in the supplementary material.
% WARN: above sentence uses `\-`.

\subsection{Qualitative Analysis}
\label{sec:qualitative}
\cref{fig:qual} presents the qualitative results. Column (a) displays the original images. Columns (b)--(f) depict the attribution maps generated by the baseline methods, overlaid on the original images. Column (g) represents the results generated by the proposed method.
Columns (b), (c), and (d) show the explanations generated by Layer-wise Relevance Propagation (LRP)~\cite{bach2015layerwiserelevancepropagationlrp}, Integrated Gradients~\cite{sundararajan2017integratedgrad}, and Guided BackPropagation~\cite{springenberg2015striving}, respectively. 
Attribution maps generated by these explanation methods were often noisy or did not sufficiently highlight relevant regions, as demonstrated by the qualitative results.
Columns (e) and (f) show the explanations generated by Grad-CAM~\cite{Selvaraju2017gradcam} and Score-CAM~\cite{wangScoreCAMScoreWeightedVisual2020}, respectively. Both of their results have attention regions that encompass the whole of the relevant objects, but also focus on the background surrounding them.
In contrast, the attribution maps generated by the proposed method specifically target the relevant objects with detailed focus and demonstrate minimal attention to background regions, thus yielding more appropriate explanations.

\begin{table*}[!t]
\renewcommand*{\arraystretch}{1.25}
\newcommand*{\bhline}[1]{\noalign{\hrule height #1}}
\caption{Quantitative results on ImageNet and the CUB dataset. IG and Guided BP denote Integrated Gradients and Guided BackPropagation, respectively. Ins. and Del. denote Insertion and Deletion score, respectively. The best results are marked in bold.}
\centering
\begin{tabular}{@{\hspace{5pt}} l @{\hspace{10pt}} ccc @{\hspace{10pt}} ccc @{\hspace{5pt}}}
\bhline{1.0pt}
\multicolumn{1}{@{\hspace{5pt}}l}{Dataset} & \multicolumn{3}{c}{CUB~\cite{WahCUB_200_2011}} & \multicolumn{3}{c}{ImageNet~\cite{deng2009Imagenet}} \\
\cline{2-4} \cline{5-7} 
\multicolumn{1}{@{\hspace{5pt}}l}{Metric [\%]} & Ins. $(\uparrow)$ & Del. $(\downarrow)$ & ID score $(\uparrow)$ & Ins. $(\uparrow)$ & Del. $(\downarrow)$ & ID score $(\uparrow)$ \\
\hline
\multicolumn{1}{@{\hspace{5pt}}l}{Method} & \multicolumn{6}{l}{}\\
\hline
LRP~\cite{bach2015layerwiserelevancepropagationlrp} & $5.8 \pm 0.2$ & $4.7 \pm 0.1$ & $1.1 \pm 0.0$ & 9.5 & 8.3 & 1.1 \\
IG~\cite{sundararajan2017integratedgrad} & $2.0 \pm 0.1$ & $1.5 \pm 0.1$ & $0.6 \pm 0.0$ & 5.2 & 6.2 & -1.1 \\
Guided BP~\cite{springenberg2015striving} & $4.2 \pm 0.2$ & $1.4 \pm 0.1$ & $2.8 \pm 0.2$ & 11.5 & 5.7 & 5.7 \\
Grad-CAM~\cite{Selvaraju2017gradcam} & $50.8 \pm 1.5$ & $5.5 \pm 0.4$ & $45.3 \pm 1.1$ & 49.7 & 12.6 & 37.1 \\
Score-CAM~\cite{wangScoreCAMScoreWeightedVisual2020} & $51.1 \pm 1.7$ & $5.4 \pm 0.4$ & $45.7 \pm 1.4$ & 48.8 & 13.3 & 35.5 \\
Ours & \scalebox{0.9}[1]{\textbf{59.5 $\pm$ 1.0}}
 & \scalebox{0.9}[1]{\textbf{1.4 $\pm$ 0.0}} & \scalebox{0.9}[1]{\textbf{58.2 $\pm$ 1.0}} & \scalebox{0.9}[1]{\textbf{56.3}} & \textbf{1.8} & \scalebox{0.9}[1]{\textbf{54.5}} \\
\bhline{1.0pt}
\end{tabular}
\label{tab:quantitative}
\end{table*}

\subsection{Quantitative Comparison Against Baselines}
\Cref{tab:quantitative} presents the quantitative results of a comparison between several baseline methods and the proposed method. For the experiments on the CUB dataset, we conducted five experiments using each method and computed the mean and standard deviation as the final results. For the experiments on the ImageNet, we conducted a single experimental run with a pretrained model.
The following methods were selected as baselines: LRP, Integrated Gradients, Guided BackPropagation, Grad-CAM, and Score-CAM.
We selected these methods (except LRP) because they are standard methods that have been successfully applied to models with skip connections.

To quantitatively evaluate our method, we employed the Insertion, Deletion, and Insertion-Deletion (ID) scores~\cite{Petsiuk2018rise,pan2021explaining,7552539}.
These are standard evaluation metrics for explanation generation tasks, and we consider the ID score as the primary evaluation metric.
The Insertion and Deletion scores were calculated as the area under the Insertion and Deletion curves, respectively. The ID score is defined as the difference between the Insertion and Deletion scores.
Please find details of these metrics in the supplementary material.
For empirical evaluation, we randomly selected 1,000 samples from the target dataset, ensuring equal representation from each class.

As listed in \Cref{tab:quantitative}, our method achieved an ID score of 0.582 in the experiments on the CUB dataset. The corresponding ID scores of LRP, Integrated Gradients, Guided BackPropagation, Grad-CAM, and Score-CAM are 0.011, 0.006, 0.028, 0.453 and 0.457, respectively. The proposed method outperformed the best baseline method, Score-CAM, by 0.125 in terms of the ID score and achieved the best performance in terms of both the Insertion and Deletion scores.
Furthermore, as listed in \Cref{tab:quantitative}, in the ImageNet experiments, our method outperformed all the baselines on the ID score. Specifically, it exceeded the highest-scoring baseline method, Grad-CAM, by 0.174 on the ID score, and again achieved the best performances in terms of both the Insertion and Deletion scores.

\subsection{Ablation Study of Relevance Propagation Rules for Bottleneck Modules}
\label{sec:ablation}

\begin{table*}[!t]
    \setlength{\tabcolsep}{4pt}
    \renewcommand*{\arraystretch}{1.25}
    \newcommand*{\bhline}[1]{\noalign{\hrule height #1}}
    \caption{Comparison of propagation rules for the Bottleneck modules. ``Include Identical'' denotes the condition in which the relevance score for skip connections with identity mapping is not set to 0. Insertion, Deletion and ID scores were calculated on ImageNet. The highest scores are marked in bold.}
    \centering
    \begin{tabular}{@{\hspace{5pt}}ccl@{\hspace{15pt}}ccc@{\hspace{5pt}}}
    \bhline{1.0pt}
    Method & \makecell{Include \\ identical} & \makecell{Relevance \\ splitting} & Insertion ($\uparrow$) & Deletion ($\downarrow$) & ID Score ($\uparrow$) \\ \hline
    (i) & & Symmetric & 0.543 & 0.033& 0.510 \\
    (ii) & $\checkmark$ & Symmetric & 0.553 & 0.036& 0.517 \\
    (iii) & & Ratio-Based & 0.543 & 0.033& 0.510 \\
    (iv) & $\checkmark$ & Ratio-Based & \textbf{0.563} & \textbf{0.018}& \textbf{0.545} \\
    \bhline{1.0pt}
    \end{tabular}
    \label{tab:ablation}
\end{table*}
\begin{figure*}[!t]
    \centering

    % ENTIRE BOX
    \begin{minipage}{0.7\linewidth}
        % Row 1
        \begin{minipage}{0.32\linewidth}
            \raisebox{4pt}{\includegraphics[width=\linewidth]{fig/qual/original/bee.png}}
        \end{minipage}
        \hfill
        \begin{minipage}{0.32\linewidth}
            \raisebox{4pt}{\includegraphics[width=\linewidth]{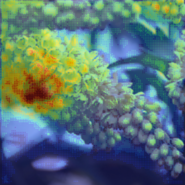}}
        \end{minipage}
        \hfill
        \begin{minipage}{0.32\linewidth}
            \raisebox{4pt}{\includegraphics[width=\linewidth]{fig/qual/ours/bee.png}}
        \end{minipage}

        % Row 2
        \begin{minipage}{0.32\linewidth}
            \raisebox{4pt}{\includegraphics[width=\linewidth]{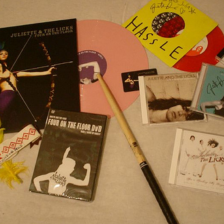}}
        \end{minipage}
        \hfill
        \begin{minipage}{0.32\linewidth}
            \raisebox{4pt}{\includegraphics[width=\linewidth]{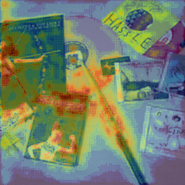}}
        \end{minipage}
        \hfill
        \begin{minipage}{0.32\linewidth}
            \raisebox{4pt}{\includegraphics[width=\linewidth]{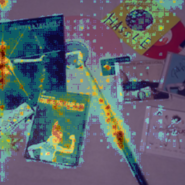}}
        \end{minipage}

        % Row 3
        \begin{minipage}[t]{0.32\linewidth}
            \centering Original
        \end{minipage}
        \hfill
        \begin{minipage}[t]{0.32\linewidth}
            \centering Method (i)
        \end{minipage}
        \hfill
        \begin{minipage}[t]{0.32\linewidth}
            \centering Method (iv)
        \end{minipage}
    \end{minipage}
    \caption{Qualitative results from the ablation study of propagation rules for the Bottleneck modules. Methods (i) and (iv) refer to the methods described in \Cref{tab:ablation}. Method \scalebox{0.975}[1]{(iv) exhibits a more concentrated attribution towards relevant objects than Method (i).}}
    \label{fig:ablation}
\end{figure*}

As discussed in \cref{sec:LRP-for-the-Bottleneck-modules,sec:LRP-for-two-types-of-skip-connections}, we formulated several types of relevance propagation rules for Bottleneck modules. This section describes the results of an ablation study focusing on the key design elements within these rules. \Cref{tab:ablation} presents the quantitative results of this ablation study.

The ID score under Method (iv) is 0.545, which surpasses the score under Method (iii) by 0.035. This indicates the effectiveness of allocating relevance to the skip connections with identity mappings. Furthermore, the ID score under Method (iv) exceeds that of Method (ii) by 0.028. This suggests that adopting Ratio-Based Splitting, which considers the element-wise proportion of the output of a skip connection and a residual connection, results in higher-quality attribution maps than those generated by Symmetric Splitting.

The performance drop from Method (iv) to Method (iii) is greater than that from Method (iv) to Method (ii). This indicates that allocating relevance to skip connections with identity mappings makes a greater contribution to the enhancement in the quality of the attribution maps.

When examining Method (iv), there is a notable reduction in the Deletion score compared with Methods (i)--(iii). This implies that Ratio-Based Splitting while also propagating relevance to skip connections with identity mappings prevents the attribution from dispersing to non-relevant objects such as the background. Indeed, as shown in \cref{fig:ablation}, when these two key design elements are used simultaneously, there is a visible reduction in the dispersion of attribution to non-relevant backgrounds compared with when neither element is used.

Furthermore, we conducted an ablation study to evaluate the Heat Quantization technique. \Cref{tab:ablation-hq} presents the quantitative results of it.
The ID score under Method (ii) is 0.510, surpassing that of Method (i) by 0.134. This indicates that Heat Quantization successfully improves the quality of the generated attribution maps.
\begin{table}[!t]
    \setlength{\tabcolsep}{4pt}
    \renewcommand*{\arraystretch}{1.25}
    \newcommand*{\bhline}[1]{\noalign{\hrule height #1}}
    \caption{\scalebox{0.97}[1]{Quantitative results of the ablation study of Heat Quantization (HQ). These} \scalebox{0.96}[1]{experimental runs were conducted on ImageNet. The best results are highlighted in bold.}}
    \centering
    \begin{tabular}{@{\hspace{5pt}}ll@{\hspace{20pt}}ccc@{\hspace{5pt}}}
    \bhline{1.0pt}
     \multicolumn{2}{c@{\hspace{20pt}}}{Method} & \multicolumn{1}{c}{Insertion ($\uparrow$)} & \multicolumn{1}{c}{Deletion ($\downarrow$)} & \multicolumn{1}{c}{ID Score ($\uparrow$)} \\ \hline
    \multicolumn{1}{c}{(i)} & w/o HQ & 0.442 & 0.066 & 0.376 \\
    \multicolumn{1}{c}{(ii)} & w/ HQ  & \textbf{0.543} & \textbf{0.033} & \textbf{0.510} \\
    \bhline{1.0pt}
    \end{tabular}
    \label{tab:ablation-hq}
\end{table}

\begin{figure*}[!t]
    \centering

    % ENTIRE BOX
    \begin{minipage}{1.0\linewidth}
        % Row 1
        \begin{minipage}{0.328\linewidth}
            \includegraphics[width=\linewidth]{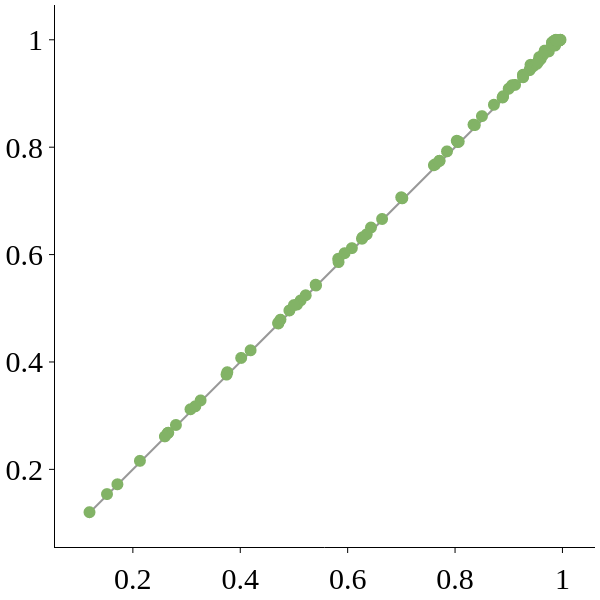}
        \end{minipage}
        \hfill
        \begin{minipage}{0.328\linewidth}
            \includegraphics[width=\linewidth]{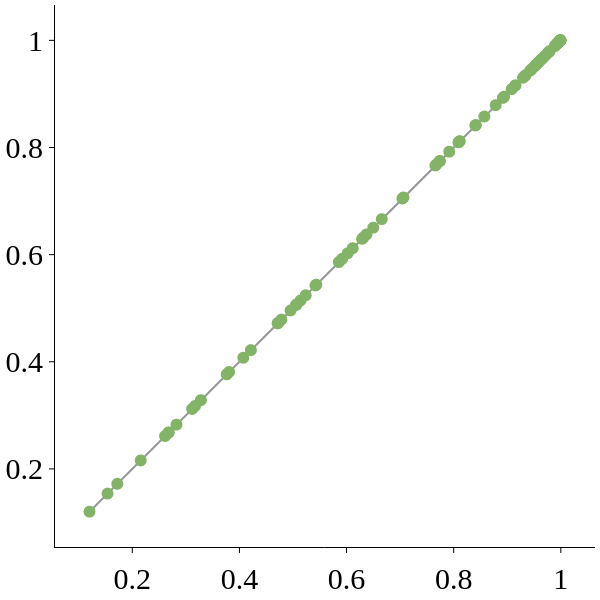}
        \end{minipage}
        \hfill
        \begin{minipage}{0.328\linewidth}
            \includegraphics[width=\linewidth]{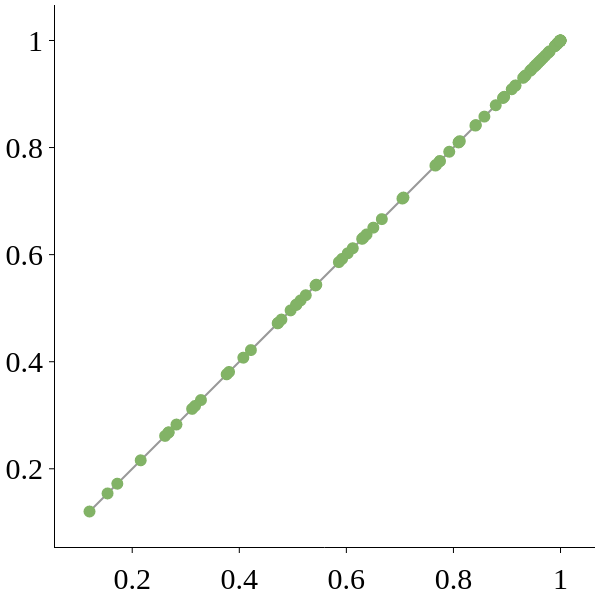}
        \end{minipage}

        % Row 2
        \begin{minipage}[t]{0.32\linewidth}
            \centering \hspace{18pt} {\footnotesize (a)}
        \end{minipage}
        \hfill
        \begin{minipage}[t]{0.32\linewidth}
            \centering \hspace{15pt} {\footnotesize (b)}
        \end{minipage}
        \hfill
        \begin{minipage}[t]{0.32\linewidth}
            \centering \hspace{12pt} {\footnotesize (c)}
        \end{minipage}
    \end{minipage}
    \caption{
    Visualization of the sum of relevance scores backpropagated to three critical checkpoints within ResNet50: (a) input of the entire network, (b) input of the first Bottleneck block, and (c) input of the last Bottleneck block.
    In each panel, the horizontal axis and the vertical axis represent the sum of relevance score at the corresponding checkpoint and model's output predictions, respectively.
    }
    \label{fig:conservation}
\end{figure*}

\subsection{Empirical Analysis of Conservation Property}
For an explanation method to be considered valid, the sum of the contributions from all inputs must equal the model's output, thereby ensuring that the entirety of the prediction's attribution is accounted for. The conservation property establishes this condition and is therefore very important in providing a rigorous foundation for the interpretability of complex models~\cite{bach2015layerwiserelevancepropagationlrp,shrikumar17deeplift,sundararajan2017integratedgrad,ali2022lrptransformer}. In this section, we show that our implementation strictly guarantees the conservation property.

We conducted an empirical analysis using 100 samples extracted from ImageNet to rigorously assess the conservation property of our method. For each sample, we visualized the sum of the relevance scores backpropagated to three checkpoints: (a) the input, (b) the input of the first Bottleneck block, and (c) the input of the last Bottleneck block.
\cref{fig:conservation} plots these sums against the model's output predictions.
The plotted points are precisely aligned with \(p(\hat y_c) = \sum_i R^{(l)}_i\), where \(p(\hat y_c)\) represents the model's predicted probability for class \(c\) and \(\sum_i R^{(l)}_i\) represents the accumulated relevance scores at each layer \(l\).
This precise alignment across all sampled points at all checkpoints
% ---(a), (b), and (c)---
demonstrates that our explanation method rigorously adheres to the conservation property.
The uniformity of these results demonstrates
our method's integrity of the relevance propagation.
\section{Conclusion}
We focused on the task of visualizing important regions in an image as a visual explanation of the model's decisions.
We extended Layer-wise Relevance Propagation (LRP)~\cite{bach2015layerwiserelevancepropagationlrp} to models with residual connections such as ResNet~\cite{he2016deep} by introducing Relevance Splitting at points where the output from a skip connection converges with that from a residual block.
Moreover, the conservation property is guaranteed throughout the proposed process, thereby preserving the integrity of the relevance propagation mechanism.
Additionally, we introduced the Heat Quantization to mitigate the issue of overconcentration of generated attributions within irrelevant regions.
Our method demonstrated superior performance to that of baseline methods on standard evaluation metrics, including the ID score.
Furthermore, we investigated improved designs for relevance propagation within residual blocks and skip connections through an ablation study on relevance propagation within the Bottleneck module and demonstrated the significance of propagating relevance through skip connections and employing Ratio-Based Relevance Splitting.

\subsubsection{Limitations and Future Directions.}
Although our experiments were conducted on ResNet, our method has the potential to be applied to various models with residual connections. Consequently, further experiments could be undertaken on other models, such as transformers~\cite{vaswani2017attention}. Additionally, our method could potentially be adapted for use in different modalities, such as natural language processing, by applying it to models specific to these modalities. We leave these further extensions for future study.

\subsubsection*{Acknowledgements}
This work was partially supported by JSPS KAKENHI Grant Number 23H03478, JST CREST, and NEDO.

% ---- Bibliography ----
%
% BibTeX users should specify bibliography style 'splncs04'.
% References will then be sorted and formatted in the correct style.
%
\bibliographystyle{splncs04}
\bibliography{main}

\begin{thebibliography}{10}
\providecommand{\url}[1]{\texttt{#1}}
\providecommand{\urlprefix}{URL }
\providecommand{\doi}[1]{https://doi.org/#1}

\bibitem{ali2022lrptransformer}
Ali, A., Schnake, T., Eberle, O., et~al.: {XAI for Transformers: Better Explanations through Conservative Propagation}. In: ICML. pp. 435--451 (2022)

\bibitem{arras2017explaininglrplstm}
Arras, L., Montavon, G., M{\"u}ller, R., et~al.: {Explaining Recurrent Neural Network Predictions in Sentiment Analysis}. In: WASSA. pp. 159--168 (2017)

\bibitem{bach2015layerwiserelevancepropagationlrp}
Bach, S., et~al.: {On Pixel-Wise Explanations for Non-Linear Classifier Decisions by Layer-Wise Relevance Propagation}. PLOS ONE  \textbf{10}(7),  1--46 (2015)

\bibitem{binderLayerwiseRelevancePropagation2016}
Binder, A., et~al.: {Layer-Wise Relevance Propagation for Neural Networks with Local Renormalization Layers}. In: ICANN. pp. 63--71 (2016)

\bibitem{cheferTransformerInterpretabilityAttention2021}
Chefer, H., Gur, S., Wolf, L.: Transformer {{Interpretability Beyond Attention Visualization}}. In: CVPR. pp. 782--791 (2021)

\bibitem{DiffusionDet}
Chen, S., Sun, P., Song, Y., Luo, P.: {DiffusionDet: Diffusion Model for Object Detection}. In: ICCV. pp. 19773--19786 (2023)

\bibitem{deng2009Imagenet}
Deng, J., Dong, W., Socher, R., Li, L.J., Li, K., Fei-Fei, L.: {ImageNet: A Large-Scale Hierarchical Image Database}. In: CVPR. pp. 248--255 (2009)

\bibitem{fong2017interpretable}
Fong, R.C., Vedaldi, A.: {Interpretable Explanations of Black Boxes by Meaningful Perturbation}. In: ICCV. pp. 3429--3437 (2017)

\bibitem{fukui2019attention}
Fukui, H., Hirakawa, T., et~al.: {Attention Branch Network: Learning of Attention Mechanism for Visual Explanation}. In: CVPR. pp. 10705--10714 (2019)

\bibitem{he2016deep}
He, K., Zhang, X., Ren, S., Sun, J.: {Deep Residual Learning for Image Recognition}. In: CVPR. pp. 770--778 (2016)

\bibitem{Hochreiter1997LSTM}
Hochreiter, S., Schmidhuber, J.: {Long Short-Term Memory}. Neural Computation  \textbf{9}(8),  1735--1780 (1997)

\bibitem{Iida2022lambdaattetnionbranchnetworkslabn}
Iida, T., Komatsu, T., Kaneda, K., et~al.: {Visual Explanation Generation Based on Lambda Attention Branch Networks}. In: ACCV. pp. 3536--3551 (2022)

\bibitem{itaya2021maska3cexplanationattentionbranch}
Itaya, H., et~al.: {Visual Explanation using Attention Mechanism in Actor-Critic-based Deep Reinforcement Learning}. In: IJCNN. pp. 1--10 (2021)

\bibitem{jacovi-etal-2023-neighboring}
Jacovi, A., Schuff, H., Adel, H., Vu, N.T., et~al.: {Neighboring Words Affect Human Interpretation of Saliency Explanations}. In: ACL. pp. 11816--11833 (2023)

\bibitem{kamath2021mdetr}
Kamath, A., Singh, M., LeCun, Y., Synnaeve, G., Misra, I., Carion, N.: {MDETR - Modulated Detection for End-to-End Multi-Modal Understanding}. In: ICCV. pp. 1780--1790 (2021)

\bibitem{krizhevsky2009learning}
Krizhevsky, A., Nair, V., Hinton, G.: {Learning Multiple Layers of Features from Tiny Images}. Tech. rep., University of Toronto (2009)

\bibitem{lundberg2017shap}
Lundberg, S., Lee, I.: {A Unified Approach to Interpreting Model Predictions}. In: NeurIPS. pp. 4765--4774 (2017)

\bibitem{aiact}
Madiaga: {Artificial Intelligence Act} (2023), \url{https://www.europarl.europa.eu/RegData/etudes/BRIE/2021/698792/EPRS_BRI(2021)698792_EN.pdf}

\bibitem{molnar2020interpretable}
Molnar, C., Casalicchio, G., et~al.: {Interpretable Machine Learning -- A Brief History, State-of-the-Art and Challenges}. In: ECML PKDD 2020 Workshops. pp. 417--431 (2020)

\bibitem{textbook}
Montavon, G., Binder, A., Lapuschkin, S., Samek, W., M{\"u}ller, K.R.: {Layer-Wise Relevance Propagation: An Overview}, pp. 193--209. Springer International Publishing (2019)

\bibitem{montavon2017dtd}
Montavon, G., Lapuschkin, S., et~al.: {Explaining Nonlinear Classification Decisions with Deep Taylor Decomposition}. Pattern Recognition  \textbf{65},  211--222 (2017)

\bibitem{OguraMSHYFK20aben}
Ogura, T., Magassouba, A., Sugiura, K., Hirakawa, T., Yamashita, T., Fujiyoshi, H., Kawai, H.: {Alleviating the Burden of Labeling: Sentence Generation by Attention Branch Encoder-Decoder Network}. RA-L  \textbf{5}(4),  5945--5952 (2020)

\bibitem{pan2021iared}
Pan, B., Panda, R., Jiang, Y., et~al.: {IA-RED$^2$: Interpretability-Aware Redundancy Reduction for Vision Transformers}. In: NeurIPS. pp. 24898--24911 (2021)

\bibitem{pan2021explaining}
Pan, D., Li, X., Zhu, D.: {Explaining Deep Neural Network Models with Adversarial Gradient Integration}. In: IJCAI (2021)

\bibitem{parekh2021framework}
Parekh, J., Mozharovskyi, P., d\textquotesingle Alch\'{e}-Buc, F.: {A Framework to Learn with Interpretation}. In: NeurIPS. pp. 24273--24285 (2021)

\bibitem{Petsiuk2018rise}
Petsiuk, V., Das, A., Saenko, K.: {RISE: Randomized Input Sampling for Explanation of Black-box Models}. In: BMVC. pp. 151--164 (2018)

\bibitem{PORWALidrid}
Porwal, P., Pachade, S., Kokare, M., et~al.: {{IDRiD: Diabetic Retinopathy}} \textendash{} {{Segmentation and Grading Challenge}}. Medical Image Analysis  \textbf{59}(101561) (2020)

\bibitem{radford2021learning}
Radford, A., Kim, J.W., Hallacy, C., Ramesh, A., Goh, G., Agarwal, S., Sastry, G., Askell, A., et~al.: {Learning Transferable Visual Models From Natural Language Supervision}. In: ICML. pp. 8748--8763 (2021)

\bibitem{A-Generalist-Agent}
Reed, S., Zolna, K., Parisotto, E., Colmenarejo, S.G., Novikov, A., Barth-maron, G., Gim{\'e}nez, M., Sulsky, Y., Kay, J., Springenberg, J.T., Eccles, T., Bruce, J., Razavi, A., et~al.: {A Generalist Agent}. TMLR  \textbf{2022} (2022)

\bibitem{ren2017faster}
Ren, S., He, K., et~al.: {Faster R-CNN: Towards Real-Time Object Detection with Region Proposal Networks}. IEEE Trans. PAMI  \textbf{39}(6),  1137--1149 (2017)

\bibitem{ribeiro2016lime}
Ribeiro, M., Singh, S., et~al.: {``Why Should I Trust You?'': Explaining the Predictions of Any Classifier}. In: KDD. pp. 1135--1144 (2016)

\bibitem{SAEED2023110273}
Saeed, W., Omlin, C.: {Explainable AI (XAI): A Systematic Meta-Survey of Current Challenges and Future Opportunities}. Knowledge-Based Systems  \textbf{263},  110273 (2023)

\bibitem{7552539}
Samek, W., Binder, A., Montavon, G., Lapuschkin, S., Müller, K.R.: {Evaluating the Visualization of What a Deep Neural Network Has Learned}. IEEE Transactions on Neural Networks and Learning Systems  \textbf{28}(11),  2660--2673 (2017)

\bibitem{Selvaraju2017gradcam}
Selvaraju, R., et~al.: {Grad-CAM: Visual Explanations from Deep Networks via Gradient-Based Localization}. In: ICCV. pp. 618--626 (2017)

\bibitem{shrikumar2016gradientxinput}
Shrikumar, A., Greenside, P., Shcherbina, A., Kundaje, A.: {Not Just a Black Box: Learning Important Features Through Propagating Activation Differences}. arXiv preprint arXiv:1605.01713  (2016)

\bibitem{shrikumar17deeplift}
Shrikumar, A., et~al.: {Learning Important Features Through Propagating Activation Differences}. In: ICML. vol.~70, pp. 3145--3153 (2017)

\bibitem{simonyan2015vgg}
Simonyan, K., Zisserman, A.: {Very Deep Convolutional Networks for Large-Scale Image Recognition}. In: ICLR. p. 1–14 (2015)

\bibitem{simonyan2014deep}
Simonyan, K., Vedaldi, A., et~al.: {Deep Inside Convolutional Networks: Visualising Image Classification Models and Saliency Maps}. In: ICLR. pp.~1--8 (2014)

\bibitem{springenberg2015striving}
Springenberg, J., Dosovitskiy, A., Brox, T., Riedmiller, M.: {Striving for Simplicity: The All Convolutional Net}. In: ICLR (workshop track) (2015)

\bibitem{srinivas2019fullgrad}
Srinivas, S., Fleuret, F.: {Full-Gradient Representation for Neural Network Visualization}. In: NeurIPS. vol.~32 (2019)

\bibitem{sundararajan2017integratedgrad}
Sundararajan, M., Taly, A., Yan, Q.: {Axiomatic Attribution for Deep Networks}. In: ICML. pp. 3319--3328 (2017)

\bibitem{vaswani2017attention}
Vaswani, A., Shazeer, N., Parmar, N., Uszkoreit, J., Jones, L., Gomez, A.N., et~al.: {Attention is All you Need}. In: NeurIPS. pp. 5998--6008 (2017)

\bibitem{WahCUB_200_2011}
Wah, C., Branson, S., et~al.: {The Caltech-UCSD Birds-200-2011 Dataset}. Tech. Rep. CNS-TR-2011-001, California Institute of Technology (2011)

\bibitem{wangScoreCAMScoreWeightedVisual2020}
Wang, H., Wang, Z., et~al.: {Score-CAM: Score-Weighted Visual Explanations for Convolutional Neural Networks}. In: CVPR. pp. 24--25 (2020)

\bibitem{NEURIPS2023_c1f7b1ed}
Wang, W., Chen, Z., Chen, X., Wu, J., Zhu, X., Zeng, G., Luo, P., Lu, T., Zhou, J., Qiao, Y., Dai, J.: {VisionLLM: Large Language Model is also an Open-Ended Decoder for Vision-Centric Tasks}. In: NeurIPS. pp. 61501--61513 (2023)

\bibitem{detic}
Zhou, X., Girdhar, R., Joulin, A., et~al.: {Detecting Twenty-Thousand Classes Using Image-Level Supervision}. In: ECCV. pp. 350--368 (2022)

\end{thebibliography}

% ---- Appendix ---
\clearpage
\renewcommand{\thesection}{\Alph{section}}
\renewcommand{\thetable}{\Alph{table}}
\renewcommand{\thefigure}{\Alph{figure}}
\setcounter{section}{0}
\setcounter{table}{0}
\setcounter{figure}{0}
\section{Additional Details of Experimental Setup}
\subsubsection{Datasets.}
We used the Caltech-UCSD Birds-200-2011 (CUB) dataset~\cite{WahCUB_200_2011} and the validation set of ImageNet~\cite{deng2009Imagenet} (ILSVRC) 2012 to evaluate our method.
These datasets were chosen because they are standard datasets for visual explanation generation tasks.
The CUB dataset contains 11,788 images from 200 classes of bird species.
The validation set of ImageNet consists of 50,000 images from 1,000 classes.

\subsubsection{Models.}
For the experiments on the CUB dataset, we employed ResNet50~\cite{he2016deep} trained on it.
To train the model, we divided the CUB dataset into training, validation, and test sets comprising 5,394, 600, and 5,794 samples, respectively. The official split of the CUB dataset provides only the training and test sets. Therefore, we used the official split for the test set and partitioned the official training set into our training and validation sets.
We used the training set to train the model and the validation set to tune the hyperparameters. We evaluated the model on the test set.
As part of the preprocessing stage, the input images were resized to a uniform dimension of 224 $\times$ 224 pixels. Moreover, during the training process, we applied flipping and cropping to the input images as data augmentation techniques.
We used a ResNet50 model pretrained on ImageNet, which is publicly available.

\Cref{tab:exp_setup} summarizes the experimental setups. We applied a learning rate decay such that the learning rate decreases by a factor of 10 every 30 epochs. We employ the cross-entropy loss as our loss function during the training of ResNet models.
In our experiments on the CUB dataset, 
We stopped the training when the loss on the validation set did not improve for six consecutive epochs.
We trained the models on a GeForce RTX 3080 with 16GB of memory and an Intel Core i9-11980HK with 64GB of memory.
It took approximately 1 h to train a ResNet50 model on the CUB dataset.
The inference time for a single sample using ResNet50 was approximately 4.66 ms. Our implementation required 40.7 ms to calculate the forward path and generate the attribution for each sample. Theoretically, the proposed explanation method has the same computational cost as backpropagation.

\begin{table}[!t]
    \setlength{\tabcolsep}{4pt}
    \renewcommand*{\arraystretch}{1.25}
    \newcommand*{\bhline}[1]{\noalign{\hrule height #1}}
    \caption{Experimental setup for the experiment on the CUB dataset.}
    \centering
    \begin{tabular}{@{\hspace{5pt}}l@{\hspace{30pt}}l@{\hspace{5pt}}}
        \bhline{1.0pt}
        Optimizer & SGD w/ momentum 0.9 \\
        Learning rate & $1.0 \times 10^{-3}$ \\
        Weight decay & $1.0 \times 10^{-4}$ \\
        \#Epoch & 300 \\
        Batch size & 32 \\
        Image size & 224 \\
        \bhline{1.0pt}
    \end{tabular}
    \label{tab:exp_setup}
\end{table}

\subsubsection{Evaluation metrics.}
To quantitatively evaluate our method, we used the Insertion, Deletion, and Insertion-Deletion (ID) scores~\cite{Petsiuk2018rise}.
The Insertion and Deletion curves represent the changes in prediction when important regions based on the final attribution $\bm \alpha$ are inserted or deleted, respectively. The details are as follows.
First, the elements of $\bm \alpha$ are sorted in descending order as $\alpha_{i_1j_1}, \alpha_{i_2j_2}, \dots, \alpha_{i_Nj_N}$ where $N$ denotes the total number of pixels. The sets $A_n$, $\bm{i}_n$, $\bm{d}_n$ are defined as:
\begin{align}
    A_n &= \{(i_k, j_k) \mid k \leq n\}, \\
    (\bm{i}_n, \bm{d}_n) &= \begin{cases} 
        (x_{ij}, 0) & \text{if } (i, j) \in A_n, \\
        (0, x_{ij}) & \text{if } (i, j) \notin A_n.
    \end{cases}
\end{align}
Here, $n$ represents the number of pixels to insert or delete.
Given inputs $\bm{i}_n$ and $\bm{d}_n$, the model outputs $\bm{y}^{(\mathrm{ins}, n)}$ and $\bm{y}^{(\mathrm{del}, n)}$, respectively.
The Insertion and Deletion curves are defined as the curves plotted for $n$ against the $c$-th elements of $\bm{y}^{(\mathrm{ins}, n)}$ and $\bm{y}^{(\mathrm{del}, n)}$, respectively, where $c$ represents the class to which $\bm x$ belongs.

\begin{table*}[!t]
\renewcommand*{\arraystretch}{1.25}
\newcommand*{\bhline}[1]{\noalign{\hrule height #1}}
\caption{Full quantitative results on ImageNet and the CUB dataset. IG and Guided BP denote Integrated Gradients and Guided BackPropagation, respectively. Ins. and Del. denote Insertion and Deletion score, respectively. The best results are marked in bold.}
\centering
\begin{tabular}{@{\hspace{5pt}} l @{\hspace{10pt}} ccc @{\hspace{10pt}} ccc @{\hspace{5pt}}}
\bhline{1.0pt}
\multicolumn{1}{@{\hspace{5pt}}l}{Dataset} & \multicolumn{3}{c}{CUB~\cite{WahCUB_200_2011}} & \multicolumn{3}{c}{ImageNet~\cite{deng2009Imagenet}} \\
\cline{2-4} \cline{5-7} 
\multicolumn{1}{@{\hspace{5pt}}l}{Metric [\%]} & Ins. $(\uparrow)$ & Del. $(\downarrow)$ & ID score $(\uparrow)$ & Ins. $(\uparrow)$ & Del. $(\downarrow)$ & ID score $(\uparrow)$ \\
\hline
\multicolumn{1}{@{\hspace{5pt}}l}{Method} & \multicolumn{6}{l}{}\\
\hline
LRP~\cite{bach2015layerwiserelevancepropagationlrp} & $5.8 \pm 0.2$ & $4.7 \pm 0.1$ & $1.1 \pm 0.0$ & 9.5 & 8.3 & 1.1 \\
IG~\cite{sundararajan2017integratedgrad} & $2.0 \pm 0.1$ & $1.5 \pm 0.1$ & $0.6 \pm 0.0$ & 5.2 & 6.2 & -1.1 \\
Guided BP~\cite{springenberg2015striving} & $4.2 \pm 0.2$ & $1.4 \pm 0.1$ & $2.8 \pm 0.2$ & 11.5 & 5.7 & 5.7 \\
DeepLIFT~\cite{shrikumar17deeplift} & $2.4 \pm 0.2$ & $1.8 \pm 0.1$ & $0.6 \pm 0.1$ & 5.6 & 6.3 & -0.7 \\
Grad-CAM~\cite{Selvaraju2017gradcam} & $50.8 \pm 1.5$ & $5.5 \pm 0.4$ & $45.3 \pm 1.1$ & 49.7 & 12.6 & 37.1 \\
Score-CAM~\cite{wangScoreCAMScoreWeightedVisual2020} & $51.1 \pm 1.7$ & $5.4 \pm 0.4$ & $45.7 \pm 1.4$ & 48.8 & 13.3 & 35.5 \\
Ours & \scalebox{0.9}[1]{\textbf{59.5 $\pm$ 1.0}}
 & \scalebox{0.9}[1]{\textbf{1.4 $\pm$ 0.0}} & \scalebox{0.9}[1]{\textbf{58.2 $\pm$ 1.0}} & \scalebox{0.9}[1]{\textbf{56.3}} & \textbf{1.8} & \scalebox{0.9}[1]{\textbf{54.5}} \\
\bhline{1.0pt}
\end{tabular}
\label{tab:full-quantitative}
\end{table*}

\section{Full Quantitative Results}
\Cref{tab:full-quantitative} presents the full quantitative comparison between the proposed method and the several baseline methods: Layer-wise Relevance Propagation (LRP)~\cite{bach2015layerwiserelevancepropagationlrp}, Integrated Gradients~\cite{sundararajan2017integratedgrad}, Guided BackPropagation~\cite{springenberg2015striving}, DeepLIFT~\cite{shrikumar17deeplift}, Grad-CAM~\cite{Selvaraju2017gradcam}, and Score-CAM~\cite{wangScoreCAMScoreWeightedVisual2020}.

As listed in \Cref{tab:full-quantitative}, our method achieved an ID score of 0.582 in the experiments on the CUB dataset.
The corresponding ID scores of LRP, Integrated Gradients, Guided BackPropagation, DeepLIFT, Grad-CAM, and Score-CAM are 0.011, 0.006, 0.028, 0.006, 0.453 and 0.457, respectively.
The proposed method outperformed the best baseline method, Score-CAM, by 0.125 in terms of the ID score and achieved the best performance in terms of both the Insertion and Deletion scores.

Furthermore, as listed in \Cref{tab:full-quantitative}, in the ImageNet experiments, our method outperformed all the baselines on the ID score. Specifically, it exceeded the highest-scoring baseline method, Grad-CAM, by 0.174 on the ID score, and again achieved the best performances in terms of both the Insertion and Deletion scores.

\begin{figure}[!t]
    \centering

    % ENTIRE BOX
    \begin{minipage}{0.80\linewidth}
        \hfill
        \begin{minipage}{0.42\linewidth}
            \centering
            \raisebox{5pt}{\includegraphics[width=\linewidth]{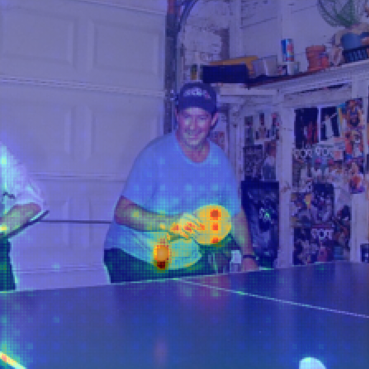}}
        \end{minipage}
        \hfill
        \begin{minipage}{0.1\linewidth}
        \end{minipage}
        \hfill
        \begin{minipage}{0.42\linewidth}
            \centering
            \raisebox{5pt}{\includegraphics[width=\linewidth]{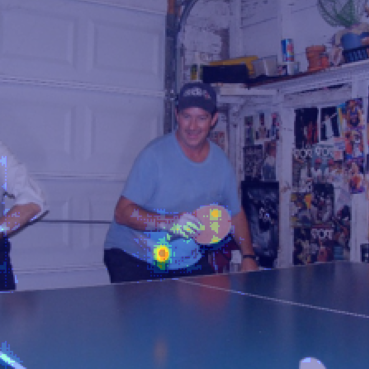}}
        \end{minipage}
        \hfill

        % Row 2
        \hfill
        \begin{minipage}[t]{0.42\linewidth}
            \centering W/o Heat Quantization
        \end{minipage}
        \hfill
        \begin{minipage}{0.1\linewidth}
        \end{minipage}
        \hfill
        \begin{minipage}[t]{0.42\linewidth}
            \centering W/ Heat Quantization
        \end{minipage}
        \hfill
    \end{minipage}
    \caption{The attribution map with respect to the ground-truth class: ``Ping Pong Ball'' without (left) and with (right) Heat Quantization.}
    \label{fig:hq}
\end{figure}

\section{Qualitative Analysis of Heat Quantization}
\cref{fig:hq} shows the qualitative result of Heat Quantization.
As shown in \cref{fig:hq}, Heat Quantization removes the excessive attribution allocated to the irrelevant regions around the ping pong ball, such as the arm and the racket, while keeping the meaningful features.

\section{Further Discussions}
\begin{figure}[!t]
    \centering

    % ENTIRE BOX
    \begin{minipage}{0.80\linewidth}
        \hfill
        \begin{minipage}{0.42\linewidth}
            \centering
            \raisebox{5pt}{\includegraphics[width=\linewidth]{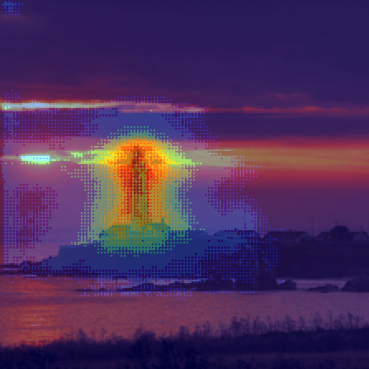}}
        \end{minipage}
        \hfill
        \begin{minipage}{0.1\linewidth}
        \end{minipage}
        \hfill
        \begin{minipage}{0.42\linewidth}
            \centering
            \raisebox{5pt}{\includegraphics[width=\linewidth]{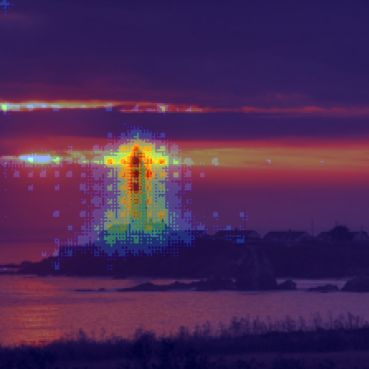}}
        \end{minipage}
        \hfill

        % Row 2
        \hfill
        \begin{minipage}[t]{0.42\linewidth}
            \centering Symmetric Splitting
        \end{minipage}
        \hfill
        \begin{minipage}{0.1\linewidth}
        \end{minipage}
        \hfill
        \begin{minipage}[t]{0.42\linewidth}
            \centering Ratio-based Splitting
        \end{minipage}
        \hfill
    \end{minipage}
    \caption{The attribution map with respect to the ground-truth class: ``Beacon'' with Symmetric Splitting (left) and Ratio-based Splitting (right).}
    \label{fig:qualitative-comparison-in-splitting}
\end{figure}

\subsection{Discussion on splitting approaches based on qualitative comparison}
We investigated the two types of splitting approaches---Symmetric Splitting and Ratio-Based Splitting---and their performance.
Through the preliminary experiments and ablation study, Ratio-Based Splitting outperforms Symmetric Splitting in ID score.
In this section, we briefly conduct a qualitative comparison between these two approaches.

As shown in \cref{fig:qualitative-comparison-in-splitting}, 
we have observed that Symmetric Splitting results in dispersed attribution. In contrast, such dispersion does not occur with Ratio-Based Splitting. This difference matches the results of quantitative comparison.
However, simultaneously, we have observed that Ratio-Based Splitting produced the grid patterns in the attribution map, though it can reduce the dispersion of attribution to non-relevant backgrounds.
We leave the theoretical validation of the relationship between the formula of both approaches and these resulting attributions
for future study.

\begin{figure}[!t]
    \centering

    % ENTIRE BOX
    \begin{minipage}{0.80\linewidth}
        \hfill
        \begin{minipage}{0.42\linewidth}
            \centering
            \raisebox{5pt}{\includegraphics[width=\linewidth]{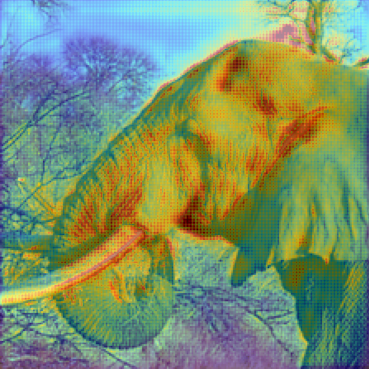}}
        \end{minipage}
        \hfill
        \begin{minipage}{0.1\linewidth}
        \end{minipage}
        \hfill
        \begin{minipage}{0.42\linewidth}
            \centering
            \raisebox{5pt}{\includegraphics[width=\linewidth]{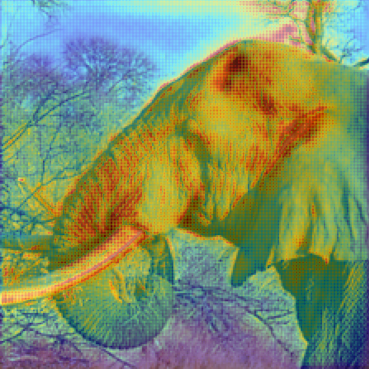}}
        \end{minipage}
        \hfill

        % Row 2
        \hfill
        \begin{minipage}[t]{0.42\linewidth}
            \centering Method (i)
        \end{minipage}
        \hfill
        \begin{minipage}{0.1\linewidth}
        \end{minipage}
        \hfill
        \begin{minipage}[t]{0.42\linewidth}
            \centering Method (ii)
        \end{minipage}
        \hfill
    \end{minipage}
    \caption{The attribution map with respect to the ground-truth class: ``African Elephant'' generated by Method (i) (left) and Method (ii) (right).}
    \label{fig:downgrade-method1-to-2}
\end{figure}

\subsection{Small performance downgrade on the deletion task from Method (i) to (ii) in Table 2}
In table 2, while Method (ii) outperforms Method (i) in the primary metric (IDscore), there is a performance downgrade from Method (i) to (ii) for the deletion task.
To elucidate the cause of this performance downgrade, we analyzed a sample in which Method (i) outperformed Method (ii) on the deletion score. As shown in \cref{fig:downgrade-method1-to-2}, in such samples, Method (ii) generates a slightly larger high-attribution region, indicated in red, than Method (i). This leads to a delayed drop in classification accuracy when higher attribution regions are deleted, thereby slightly downgrading the deletion score.

\begin{table}[!t]
    \setlength{\tabcolsep}{4pt}
    \renewcommand*{\arraystretch}{1.25}
    \newcommand*{\bhline}[1]{\noalign{\hrule height #1}}
    \caption{Comparison of different LRP rules on the ImageNet. Ins. and Del. refer to Insertion and Deletion scores, respectively.}
    \centering
    \begin{tabular}{lccc}
    \bhline{1.0pt}
    Method [\%] & \multicolumn{1}{c}{Ins. ($\uparrow$)} & \multicolumn{1}{c}{Del. ($\downarrow$)} & \multicolumn{1}{c}{ID Score ($\uparrow$)} \\ \hline
    LRP ($\epsilon$ rule) & 9.5 & 8.3 & 1.1 \\
    LRP (Mixture of rules)  & 8.6 & 7.5 & 1.1 \\
    Ours (Mixture of rules) & 64.6 & 0.2 & 64.4 \\
    Ours ($z^+$ rule) & 56.3 & 1.8 & 54.5 \\
    \bhline{1.0pt}
    \end{tabular}
    \label{tab:rules}
\end{table}

\subsection{Discussion on the LRP rules}
We used the \(\epsilon\)~rule to obtain the baseline results. To investigate the impact of the choice of LRP rules for the baseline results, we conducted additional experiments using the mixture of rules. The mixture of rules employs the \(\epsilon\)~rule for layers after the eighth bottleneck block and the \(\alpha1\beta0\)~rule for the other layers.
The \(\alpha1\beta0\) rule is equivalent to the \(z+\)~rule~\cite{textbook}, which we have adopted in our proposed method.
As shown in \Cref{tab:rules}, we observed that our mixture of rules performs comparably to the \(\epsilon\)~rule.

Furthermore, to examine the impact of LRP rule selection on our proposed method, 
we applied the same mixture of rules to our approach. As shown in \Cref{tab:rules}, we did not observe any critical influence on the results by changing the rules; rather, the quantitative results slightly improved.

\subsection{Error Analysis}
We conducted an error analysis on ImageNet. We defined a failure of explanation generation as cases where the ID score was not greater than 0.371, the highest score achieved by the baseline methods (specifically by Grad-CAM~\cite{Selvaraju2017gradcam}). There were 311 instances in which the explanation generation process failed. These failures represent 31.1\% of the total generated attribution maps.

\begin{figure*}[!t]
    \centering

    % ENTIRE BOX
    \begin{minipage}{1.0\linewidth}
        % Row 1
        \begin{minipage}{0.32\linewidth}
            \raisebox{3pt}{\includegraphics[width=\linewidth]{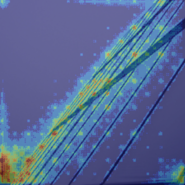}}
        \end{minipage}
        \hfill
        \begin{minipage}{0.32\linewidth}
            \raisebox{3pt}{\includegraphics[width=\linewidth]{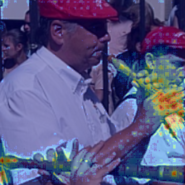}}
        \end{minipage}
        \hfill
        \begin{minipage}{0.32\linewidth}
            \raisebox{3pt}{\includegraphics[width=\linewidth]{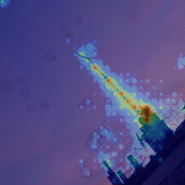}}
        \end{minipage}

        % Row 2
        \begin{minipage}[t]{0.32\linewidth}
            \centering ``Solar Collector''
        \end{minipage}
        \hfill
        \begin{minipage}[t]{0.32\linewidth}
            \centering ``Oboe''
        \end{minipage}
        \hfill
        \begin{minipage}[t]{0.32\linewidth}
            \centering ``Bubble''
        \end{minipage}

        % Row 3
        \begin{minipage}[t]{0.32\linewidth}
            \centering (a) Insufficiently Attended
        \end{minipage}
        \hfill
        \begin{minipage}[t]{0.32\linewidth}
            \centering (b) Over-Attended
        \end{minipage}
        \hfill
        \begin{minipage}[t]{0.32\linewidth}
            \centering (c) Wrongly Attended
        \end{minipage}
    \end{minipage}
    \caption{Examples of failure categories in relevance attribution. (a) Insufficiently Attended where the area of focus is too narrow and fails to cover all relevant regions. (b) Over-Attended where an excessively large area is deemed relevant, indicating a lack of precision in identifying relevant features. (c) Wrongly Attended where the relevance is attributed to areas that do not directly contribute to the classification decision.}
    \label{fig:errors}
\end{figure*}

For the error analysis, we investigated 100 samples selected in ascending order based on their ID scores. The causes of failure in these samples can be broadly categorized into three groups:
\begin{itemize}
    \item [$\bullet$] IA (Insufficiently Attended): This category includes cases where the area of attribution is too small, indicating insufficient focus on relevant regions. \cref{fig:errors} (a) illustrates an IA example.
    \item [$\bullet$] OA (Over-Attended): This category represents cases where the area of relevance is excessively large, suggesting that too much of the image is being considered relevant. \cref{fig:errors} (b) presents an OA example.
    \item [$\bullet$] WA (Wrongly Attended): This category comprises cases where relevance is assigned to pixels that do not directly contribute to the classification. An example of WA is shown in \cref{fig:errors} (c).
\end{itemize}

\begin{table}[!t]
    \setlength{\tabcolsep}{8pt}
    \renewcommand*{\arraystretch}{1.3}
    \newcommand*{\bhline}[1]{\noalign{\hrule height #1}}
    \caption{Number of failure cases in each failure category. IA, OA and WA are the three failure categories: Insufficiently Attended, Over-Attended, and Wrongly Attended, respectively. The number of the most frequent failures is highlighted in bold.}
    \centering
    \begin{tabular}{@{\hspace{5pt}}l@{\hspace{25pt}}ccc@{\hspace{5pt}}}
        \bhline{1.0pt}
        Failure category & IA & OA & WA \\ \hline
        \#Failure & \textbf{40} & 25 & 35 \\
        \bhline{1.0pt}
    \end{tabular}
    \label{tab:errors}
\end{table}

\Cref{tab:errors} indicates that the most common cause of errors is focusing on an insufficient area of the image. This observation suggests that our method might, in some instances, excessively restrict the distribution of attribution. This finding is consistent with our method’s tendency to minimize the dispersion of attribution to non-relevant regions,
as observed in \cref{sec:ablation} in the main text,
which could contribute to this issue. This highlights a potential area for refinement in our relevance score splitting rule. Adjusting the splitting rule accordingly could reduce the number of errors.

\begin{figure*}[!t]
    \centering
    \begin{minipage}[t]{0.90\linewidth}
        \begin{minipage}[t]{0.05\linewidth}
            \rotatebox[origin=c]{90}{Original}
        \end{minipage}
        \hfill
        \begin{minipage}{0.30\linewidth}
            \raisebox{5pt}{\includegraphics[width=\linewidth]{fig/qual/original/water_ouzel.png}}
        \end{minipage}
        \hfill
        \begin{minipage}{0.30\linewidth}
            \raisebox{5pt}{\includegraphics[width=\linewidth]{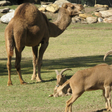}}
        \end{minipage}
        \hfill
        \begin{minipage}{0.30\linewidth}
            \raisebox{5pt}{\includegraphics[width=\linewidth]{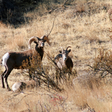}}
        \end{minipage}

        \begin{minipage}[t]{0.05\linewidth}
            \rotatebox[origin=c]{90}{LRP~\cite{bach2015layerwiserelevancepropagationlrp}}
        \end{minipage}
        \hfill
        \begin{minipage}{0.30\linewidth}
            \raisebox{5pt}{\includegraphics[width=\linewidth]{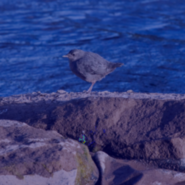}}
        \end{minipage}
        \hfill
        \begin{minipage}{0.30\linewidth}
            \raisebox{5pt}{\includegraphics[width=\linewidth]{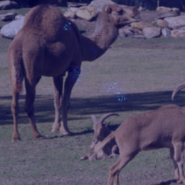}}
        \end{minipage}
        \hfill
        \begin{minipage}{0.30\linewidth}
            \raisebox{5pt}{\includegraphics[width=\linewidth]{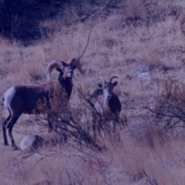}}
        \end{minipage}

        \begin{minipage}[t]{0.05\linewidth}
            \rotatebox[origin=c]{90}{Grad-CAM~\cite{Selvaraju2017gradcam}}
        \end{minipage}
        \hfill
        \begin{minipage}{0.30\linewidth}
            \raisebox{5pt}{\includegraphics[width=\linewidth]{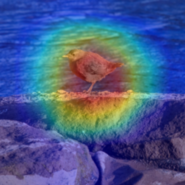}}
        \end{minipage}
        \hfill
        \begin{minipage}{0.30\linewidth}
            \raisebox{5pt}{\includegraphics[width=\linewidth]{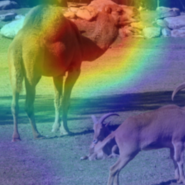}}
        \end{minipage}
        \hfill
        \begin{minipage}{0.30\linewidth}
            \raisebox{5pt}{\includegraphics[width=\linewidth]{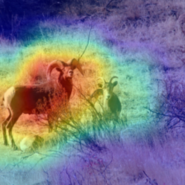}}
        \end{minipage}

        \begin{minipage}[t]{0.05\linewidth}
            \rotatebox[origin=c]{90}{Score-CAM~\cite{wangScoreCAMScoreWeightedVisual2020}}
        \end{minipage}
        \hfill
        \begin{minipage}{0.30\linewidth}
            \raisebox{5pt}{\includegraphics[width=\linewidth]{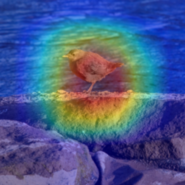}}
        \end{minipage}
        \hfill
        \begin{minipage}{0.30\linewidth}
            \raisebox{5pt}{\includegraphics[width=\linewidth]{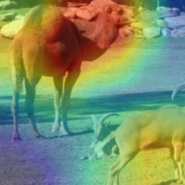}}
        \end{minipage}
        \hfill
        \begin{minipage}{0.30\linewidth}
            \raisebox{5pt}{\includegraphics[width=\linewidth]{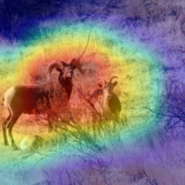}}
        \end{minipage}

        \begin{minipage}[t]{0.05\linewidth}
            \rotatebox[origin=c]{90}{Ours}
        \end{minipage}
        \hfill
        \begin{minipage}{0.30\linewidth}
            \raisebox{5pt}{\includegraphics[width=\linewidth]{fig/qual/ours/water_ouzel.png}}
        \end{minipage}
        \hfill
        \begin{minipage}{0.30\linewidth}
            \raisebox{5pt}{\includegraphics[width=\linewidth]{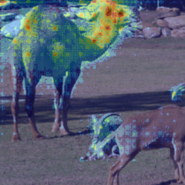}}
        \end{minipage}
        \hfill
        \begin{minipage}{0.30\linewidth}
            \raisebox{5pt}{\includegraphics[width=\linewidth]{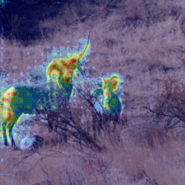}}
        \end{minipage}
    \end{minipage}

    \caption{Attribution maps produced by each explanation method for the prediction of ResNet50 with respect to the ground-truth classes (left to right):
    ``Water Ouzel,''
    ``Arabian Camel,''
    ``Ram.''
    }
    \label{fig:qual-full-1}
\end{figure*}

\begin{figure*}[!t]
    \centering
    \begin{minipage}[t]{0.90\linewidth}
        \begin{minipage}[t]{0.05\linewidth}
            \rotatebox[origin=c]{90}{Original}
        \end{minipage}
        \hfill
        \begin{minipage}{0.30\linewidth}
            \raisebox{5pt}{\includegraphics[width=\linewidth]{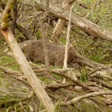}}
        \end{minipage}
        \hfill
        \begin{minipage}{0.30\linewidth}
            \raisebox{5pt}{\includegraphics[width=\linewidth]{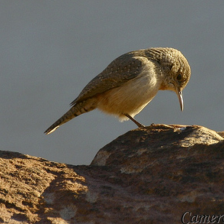}}
        \end{minipage}
        \hfill
        \begin{minipage}{0.30\linewidth}
            \raisebox{5pt}{\includegraphics[width=\linewidth]{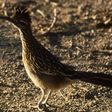}}
        \end{minipage}

        \begin{minipage}[t]{0.05\linewidth}
            \rotatebox[origin=c]{90}{LRP~\cite{bach2015layerwiserelevancepropagationlrp}}
        \end{minipage}
        \hfill
        \begin{minipage}{0.30\linewidth}
            \raisebox{5pt}{\includegraphics[width=\linewidth]{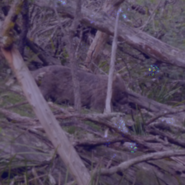}}
        \end{minipage}
        \hfill
        \begin{minipage}{0.30\linewidth}
            \raisebox{5pt}{\includegraphics[width=\linewidth]{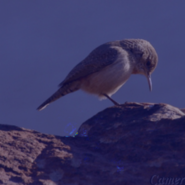}}
        \end{minipage}
        \hfill
        \begin{minipage}{0.30\linewidth}
            \raisebox{5pt}{\includegraphics[width=\linewidth]{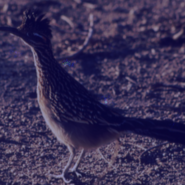}}
        \end{minipage}

        \begin{minipage}[t]{0.05\linewidth}
            \rotatebox[origin=c]{90}{Grad-CAM~\cite{Selvaraju2017gradcam}}
        \end{minipage}
        \hfill
        \begin{minipage}{0.30\linewidth}
            \raisebox{5pt}{\includegraphics[width=\linewidth]{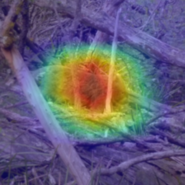}}
        \end{minipage}
        \hfill
        \begin{minipage}{0.30\linewidth}
            \raisebox{5pt}{\includegraphics[width=\linewidth]{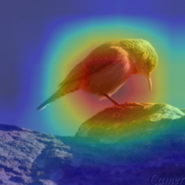}}
        \end{minipage}
        \hfill
        \begin{minipage}{0.30\linewidth}
            \raisebox{5pt}{\includegraphics[width=\linewidth]{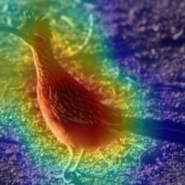}}
        \end{minipage}

        \begin{minipage}[t]{0.05\linewidth}
            \rotatebox[origin=c]{90}{Score-CAM~\cite{wangScoreCAMScoreWeightedVisual2020}}
        \end{minipage}
        \hfill
        \begin{minipage}{0.30\linewidth}
            \raisebox{5pt}{\includegraphics[width=\linewidth]{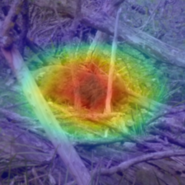}}
        \end{minipage}
        \hfill
        \begin{minipage}{0.30\linewidth}
            \raisebox{5pt}{\includegraphics[width=\linewidth]{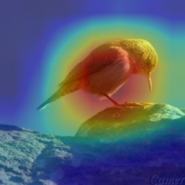}}
        \end{minipage}
        \hfill
        \begin{minipage}{0.30\linewidth}
            \raisebox{5pt}{\includegraphics[width=\linewidth]{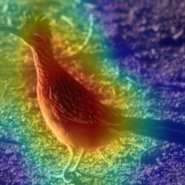}}
        \end{minipage}

        \begin{minipage}[t]{0.05\linewidth}
            \rotatebox[origin=c]{90}{Ours}
        \end{minipage}
        \hfill
        \begin{minipage}{0.30\linewidth}
            \raisebox{5pt}{\includegraphics[width=\linewidth]{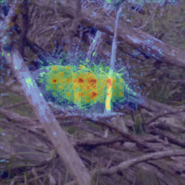}}
        \end{minipage}
        \hfill
        \begin{minipage}{0.30\linewidth}
            \raisebox{5pt}{\includegraphics[width=\linewidth]{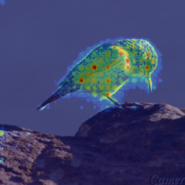}}
        \end{minipage}
        \hfill
        \begin{minipage}{0.30\linewidth}
            \raisebox{5pt}{\includegraphics[width=\linewidth]{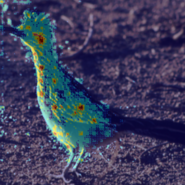}}
        \end{minipage}
    \end{minipage}
    \caption{Attribution maps produced by each explanation method for the prediction of ResNet50 with respect to the ground-truth classes (left to right):
    ``Wombat,''
    ``Rock Wren,''
    ``Geococcyx.''
    }
    \label{fig:qual-full-2}
\end{figure*}

\section{Additional Qualitative Results}
 \cref{fig:qual-full-1,fig:qual-full-2} show additional qualitative comparisons of our proposed method against baseline methods. In order to provide a clearer view, the panels show larger images than those presented in the main text.
In each figure, the top row displays the original image, while the bottom row presents the attribution maps generated by our method. The rows in the middle show the results from baseline methods.

Attribution maps generated by the original Layer-wise Relevance Propagation (LRP)~\cite{bach2015layerwiserelevancepropagationlrp} tended to be noisy and often failed to sufficiently highlight relevant regions, as demonstrated in the second rows of \cref{fig:qual-full-1,fig:qual-full-2}.
The third and fourth rows of \cref{fig:qual-full-1,fig:qual-full-2} present explanations generated by Grad-CAM~\cite{Selvaraju2017gradcam} and Score-CAM~\cite{wangScoreCAMScoreWeightedVisual2020}, respectively. Results from both methods have attention regions that encompass the whole of the relevant objects, but also include the background surrounding them.
In contrast, attribution maps generated by our proposed method sharply focus on the relevant objects and demonstrate minimal attention to irrelevant regions such as background, thus yielding more appropriate explanations.

\end{document}